\documentclass[conf]{new-aiaa}
\usepackage[utf8]{inputenc}

\usepackage{cleveref}
\usepackage{graphicx}
\usepackage{amsmath}
\usepackage[version=4]{mhchem}
\usepackage{siunitx}
\usepackage{longtable,tabularx}
\usepackage{glossaries}
\usepackage{algorithm}
\usepackage{algpseudocode}
\setkeys{glslink}{hyper=false}
\makeglossaries

\crefformat{section}{Section~#2#1#3}
\crefformat{fig}{Figure~#2#1#3}

\newcommand{\kron}{\otimes}
\newcommand{\real}{\mathbb{R}}
\newcommand{\norm}[1]{\left\lVert\, #1 \,\right\rVert}

\newcommand{\E}[1]{\mathbb{E}\left[#1\right]}

\newcommand{\Cov}[1]{\mathbb{C}\text{ov}\left[#1\right]}
\newcommand{\Tr}[1]{\text{Tr}\left(#1\right)}

\DeclareMathOperator*{\argmin}{arg\,min}
\DeclareMathOperator*{\argmax}{arg\,max}

\newsavebox{\foobox}
\newcommand{\slantbox}[2][0]{\mbox{%
        \sbox{\foobox}{#2}%
        \hskip\wd\foobox
        \pdfsave
        \pdfsetmatrix{1 0 #1 1}%
        \llap{\usebox{\foobox}}%
        \pdfrestore
}}
\newcommand\unslant[2][-.25]{\slantbox[#1]{$#2$}}

\newcommand{\cvec}{\mathbf{c}}
\newcommand{\dvec}{\mathbf{d}}

\newcommand{\fvec}{\mathbf{f}}
\newcommand{\gvec}{\mathbf{g}}

\newcommand{\ovec}{\mathbf{o}}

\newcommand{\rvec}{\mathbf{r}}

\newcommand{\uvec}{\mathbf{u}}

\newcommand{\xvec}{\mathbf{x}}
\newcommand{\yvec}{\mathbf{y}}
\newcommand{\zvec}{\mathbf{z}}

\newcommand{\Amat}{\mathbf{A}}

\newcommand{\Cmat}{\mathbf{C}}
\newcommand{\Dmat}{\mathbf{D}}

\newcommand{\Hmat}{\mathbf{H}}

\newcommand{\Omat}{\mathbf{O}}
\newcommand{\Pmat}{\mathbf{P}}

\newcommand{\Umat}{\mathbf{U}}

\newcommand{\Xmat}{\mathbf{X}}
\newcommand{\Ymat}{\mathbf{Y}}

\newcommand{\Sigmamat}{\mathbf{\Sigma}}

\newcommand{\chat}{\hat{\cvec}}

\newcommand{\fhat}{\hat{\fvec}}

\newcommand{\ohat}{\hat{\ovec}}
\newcommand{\xhat}{\hat{\xvec}}

\newcommand{\Ahat}{\hat{\Amat}}

\newcommand{\Chat}{\hat{\Cmat}}

\newcommand{\Hhat}{\hat{\Hmat}}

\newcommand{\Ohat}{\hat{\Omat}}

\newcommand{\Xhat}{\hat{\Xmat}}

\newcommand{\xdot}{\dot{\xvec}}

\newcommand{\Xdot}{\dot{\Xmat}}

\newcommand{\xhatdot}{\dot{\xhat}}

\newcommand{\Xhatdot}{\dot{\Xhat}}

\title{Multifidelity Operator Inference for Non-intrusive Model Reduction under Data Scarcity}

\author{Vivian Zhang\footnote{Undergraduate Student, School of Aerospace Engineering, AIAA Student Member, vzhang40@gatech.edu},  Tomoki Koike\footnote{PhD Candidate, School of Aerospace Engineering, AIAA Student Member, tkoike@gatech.edu}, Atticus Rex\footnote{PhD Candidate, School of Aerospace Engineering, arex8@gatech.edu}, and Elizabeth Qian\footnote{Assistant Professor, Schools of Aerospace Engineering \& Computational Science and Engineering, AIAA Member, eqian@gatech.edu}}
\affil{Georgia Institute of Technology, Atlanta, Georgia, 30332}

\newacronym{ode}{ODE}{ordinary differential equation}
\newacronym{pod}{POD}{proper orthogonal decomposition} 
\newacronym{svd}{SVD}{singular value decomposition}
\newacronym{opinf}{OpInf}{Operator Inference}
\newacronym{mfopinf}{MF-OpInf}{Multifidelity Operator Inference}
\newacronym{iid}{i.i.d.}{independently and identically distributed}
\newacronym{pde}{PDE}{partial differential equation}
\newacronym{grf}{GRF}{Gaussian Random Field}
\newacronym{kl}{KL}{Karhunen-Lo\`eve}
\newacronym{flops}{FLOPs}{floating point operations}
\newacronym{mfmc}{MFMC}{Multifidelity Monte Carlo}
\newacronym{ml}{ML}{Machine Learning}
\newacronym{nn}{NN}{neural networks}
\newacronym{dmd}{DMD}{dynamic mode decomposition}
\newacronym{gpr}{GPR}{Gaussian process regression}
\newacronym{kse}{KSE}{Kuramoto-Sivashinsky equation}

\begin{document}

\maketitle

\begin{abstract}
Many-query analyses, in which a computational model is repeatedly evaluated, play an important role in uncertainty quantification and optimization for engineering design. However, many-query analyses can be prohibitively expensive when considering high-fidelity numerical models, especially those governed by partial differential equations (PDEs). Surrogate modeling replaces costly simulations with cheap approximations, enabling practical uncertainty quantification and design workflows. Machine learning (ML) approaches to surrogate modeling have gained significant traction in the development of surrogate models due to their non-intrusive nature; they do not require access to the underlying simulation code. The Operator Inference (OpInf) approach, in particular, fits reduced operators to state and corresponding time derivative data. However, high-fidelity training data is scarce due to limited computational budgets, and OpInf models trained on such scarce data often exhibit high variance, a critical liability in high-risk engineering design. We propose Multifidelity Operator Inference (MF-OpInf), which reduces this variance by supplementing scarce high-fidelity snapshots with abundant low-fidelity data drawn from coarser discretizations of the same problem. MF-OpInf handles data across varying fidelity levels by utilizing a multifidelity linear regression framework to estimate the reduced operators. Numerical experiments on model PDE problems demonstrate that MF-OpInf predicts PDE solution fields with higher accuracy compared to the high-fidelity-only OpInf with the same computational budget.
\end{abstract}

\section{Nomenclature}

{\renewcommand\arraystretch{1.0} 
\begin{longtable*}{@{}l @{\quad=\quad} l@{}}
$N$ & full model state dimension\\
$r$ & reduced model state dimension\\
$\mathbf{\Phi}_r$ & POD basis in $\real^{N \times r}$\\
$\xvec, \xhat$ & full and reduced state in $\real^N$ and $\real^r$\\
$\Amat, \Ahat$ & full and reduced linear operators in $\real^{N \times N}$ and $\real^{r \times r}$ \\
$\Hmat, \Hhat$ & full and reduced quadratic operators in $\real^{N \times N^2}$ and $\real^{r \times r^2}$ \\
$K$ & number of snapshots \\
$\Xmat_K, \Xhat_K$ & full and reduced state snapshot matrix in $\real^{N \times K}$ \\
$\Xdot_K, \Xhatdot_K$ & full and reduced state time derivative snapshot matrix in $\real^{N \times K}$ and $\real^{r \times K}$ \\
$\Xhat^\kron_K$ & reduced quadratic state snapshot matrix in $\real^{r^2 \times K}$\\
$\Dmat_K$  & data matrix from concatenated reduced states in $\real^{(r + r^2) \times K}$ \\
$\Omat$ & reduced operator matrix in $\real^{(r + r^2)\times r}$\\
$\zvec$ & linear regression input in $\real^r$\\
$\fvec^{(1)}, \fvec^{(2)}$ & high and low-fidelity mapping from reduced initial condition to reduced initial time derivative\\
$\yvec^{(1)}, \yvec^{(2)}$ & high and low-fidelity reduced initial time derivative in $\real^r$ and $\real^r$ \\
$\Ymat_K^{(1)}, \Ymat_K^{(2)}$ & high and low-fidelity reduced initial time derivative snapshot matrices in $\real^{r \times K}$ and $\real^{r \times K}$ \\
$\tilde{\dvec}(\zvec)$ & linear regression feature mapping acting on input $\zvec$\\
$K_{1}, K_{2}$ & number of high and low-fidelity snapshots\\
$\alpha_i$ & scalar control variate coefficient for estimation of the $i$-th reduced operator column\\
$p$ & computational budget\\
$w_1, w_2$ & high and low-fidelity model costs\\
$\mu$ & partial differential equation parameter\\
$(\eta, t)$ & spatial variable and time variable\\
$\kron$ & Kronecker product\\
$\norm{\cdot}_2, \norm{\cdot}_\text{F}$ & Euclidean and Frobenius norm\\
$\E{\cdot}, \Cov{\cdot, \cdot}$ & expected value, covariance\\

\end{longtable*}}

\section{Introduction}
\lettrine{S}cientists and engineers rely on many-query analyses for decision-making, where they utilize computational models to simulate a system of interest many times with different inputs, initializations, or parameters. Examples of many-query analyses include design optimization~\cite{design_optimization}, uncertainty quantification~\cite{UQ_chapter}, data assimilation~\cite{data_assimilation}, and real-time control~\cite{optimal_control, cyberphysical_system}. However, repeated evaluations of these scientific and engineering models are computationally expensive as they generally require numerical solutions of large-scale \glspl{pde}. We refer to accurate, but computationally expensive, simulations as \emph{high-fidelity} models. This computational barrier necessitates the development and use of \textit{surrogate models}, a cheaper approximation of the high-fidelity model of interest. 

\Gls{ml} methods are one attractive way to learn surrogate models because they are \emph{non-intrusive}, allowing us to view engineering models as black-boxes, bypassing the need for access to the underlying code. Non-intrusive methods are attractive when dealing with confidential software where the underlying mechanics are hidden or with legacy code. In these cases, high-fidelity data is available, but not the model itself. \Gls{nn} in particular can learn nonlinear relationships between data and have been widely adopted in scientific and engineering applications~\cite{nn_engineering,deeponet,fnos}, with successful applications in airfoil design optimization~\cite{airfoildeepneuralop}, thermal modeling~\cite{nn_thermal}, and metamaterial design~\cite{nn_metamaterial}. Neural networks provide extremely flexible functional approximations. In contrast, the \Gls{opinf} method~\cite{peherstorfer2016data} learns a more structured \gls{ml} approximation for dynamical systems based on projection-based model reduction~\cite{model_reduction_survey, largescaledynamics}. \Gls{opinf} fits polynomial reduced operators to reduced data. 
More general nonlinearities can be handled via the use of variable transformations~\cite{qian_lift_2020}. Another \gls{ml} and model reduction approach for dynamical systems is the \gls{dmd} which fits a reduced linear evolution operator to reduced snapshot data~\cite{dmd, dmd_data}. An alternative \gls{ml} approach to learn system dynamics is to learn the coefficients of the governing equations through sparse regression of snapshot data~\cite{SINDy}.

However, a critical challenge in using these methods for practical settings is that training data are limited. Previous \gls{ml} approaches assume that high-fidelity data from a single source are readily available, but in scientific and engineering settings, generating such datasets require expensive, repeated simulation of a computational model. Training models on \emph{scarce} data typically yields models with high-variance and poor extrapolation. This motivates the development of \gls{ml} surrogate methods that achieve high prediction accuracy while compensating for the scarce high-fidelity data.

To address challenges arising from data scarcity, we consider the multifidelity setting, in which engineers have access not only to the expensive high-fidelity model but also one or more lower-fidelity models that are less accurate but cheaper to evaluate. These low-fidelity models could be simplified models under certain assumptions, a coarsened grid, or even other surrogate models from the high-fidelity simulation. Multifidelity methods seek to leverage abundant low-fidelity data to account for scarce high-fidelity data for computational acceleration and accuracy. Multifidelity methods have been successfully used for optimization~\cite{mf_optimization}, uncertainty quantification~\cite{mf_uq}, and inference~\cite{mf_bi}.

For multifidelity \gls{ml}, extensions of \gls{gpr}, where surrogate models are selected from a Gaussian distribution of functions~\cite{williams2006gaussian}, have been developed through multifidelity \gls{gpr}~\cite{mf_gp} and cokriging methods~\cite{kennedyohagan}. In these approaches, model parameters are fit to multifidelity data by exploiting the correlation structure between high and low-fidelity datasets, with extensions that build off the general \gls{gpr} model~\cite{kennedyohagan, atti_magpi, perdikaris_multifidelity, raissi_deep_2016}. Additionally, multifidelity extensions to neural operator learning approaches have been developed~\cite{mf_fno,mf_deeponet}.
Multifidelity studies have also been conducted in transfer learning, where models are pretrained and updated using datasets of varying fidelities and have been tested using both experimental and computational data~\cite{mf_cfd} as well as data with varying discretization~\cite{mf_autoencoder}. General multifidelity frameworks have also been developed for Bayesian learning approaches~\cite{mf_bayes}.

In this work, we use multifidelity control variate methods for reduced operator learning. Multifidelity control variates build off of approximate control variate estimation theory~\cite{general_ACV} in which the goal is to decrease the variance from the standard Monte Carlo estimator~\cite{MonteCarlo}. At its core, the multifidelity control variate framework uses a Monte Carlo estimate of a low number of high-fidelity samples and a correction term made up of a large number of low-fidelity estimates. The multifidelity control variate term is formulated in a way to be unbiased to high-fidelity statistics. Multifidelity control variate methods include multifidelity Monte Carlo (MFMC)~\cite{mfmc_UQ, mfmc_modelmanage}, multilevel Monte Carlo (MLMC)~\cite{mlmc}, Multilevel Best Linear Unbiased Estimators (MLBLUE)~\cite{mlblue}, and Extended Multifidelity Monte Carlo \cite{rex2026analyticallyoptimalsampleallocations}. These methods differ in the types of low-fidelity models and structure of the low-fidelity correction term. In the context of \gls{ml}, multifidelity control variate methods have been applied to the linear regression~\cite{qian2025multifidelity} setting, where the MFMC framework is used to estimate model parameters, demonstrating the potential for generalizing control variate methods for multifidelity machine learning. Our goal is to extend multifidelity machine learning for learning reduced models for dynamical systems with \gls{opinf}.

In this work, we present a novel multifidelity control variate training strategy for the \gls{opinf} reduced model learning approach, which is based on fitting reduced operators to dynamical systems data via linear regression. Our contributions are:
\begin{enumerate}
    \item We provide a new method for obtaining reduced operators from high- and low-fidelity data via multifidelity control variates. 
    \item We provide three numerical PDE examples which show that models learned from scarce high-fidelity data using the multifidelity approach are able to predict solution fields with a lower error compared using single, high-fidelity data with comparable computational cost.
\end{enumerate}

The remainder of this paper is organized as follows.~\Cref{section: background} provides background on \gls{opinf} and summarizes the multifidelity linear regression framework with the \gls{mfmc} approach.~\Cref{section: mfopinf} presents the new multifidelity approach to \gls{opinf} and provides a discussion on hyperparameter selection.~\Cref{section: numerics} quantifies the performance of \gls{mfopinf} with three numerical examples, and~\Cref{section: conclude} concludes our paper and summarizes key results and future work.

\section{Background} \label{section: background}

In this section, we introduce the proper orthogonal decomposition for model reduction (\Cref{subsection: POD}), non-intrusive \gls{opinf} method for learning reduced models (\Cref{subsection: opinf}) and the multifidelity linear regression framework (\Cref{subsection: mflr}).

\subsection{Proper Orthogonal Decomposition} \label{subsection: POD}
\Gls{pod} is a widely used model reduction method that originated in the study of turbulent flows~\cite{berkooz1993proper}.  We present POD for the following $N$-dimensional linear-quadratic \gls{ode} 
\begin{equation} \label{eq: full_lq_model}
    \xdot(t) = \Amat \xvec(t) + \Hmat (\xvec(t) \kron \xvec(t)), \qquad \xvec(0) = \xvec_0,
\end{equation}
where $\xvec (t) \in \real^N$ is the state with initial condition $\xvec_0$, and $\kron$ denotes the Kronecker product for vectors. The matrices $\Amat \in \real^{N \times N}$ and $\Hmat \in \real^{N \times N^2}$ are the linear and quadratic operators, respectively. We consider the case where the model \eqref{eq: full_lq_model} is the result of a high-dimensional discretization of a \gls{pde}, so that $N$ is large and simulating \eqref{eq: full_lq_model} is computationally expensive.

The definition of the POD basis relies on the availability of state snapshot data. With \( \xvec_i = \xvec(t_i) \), let $\xvec_1, \xvec_2, \dots, \xvec_K$ denote a collection of $K$ state snapshots which can be organized in the following snapshot matrix:  
\begin{equation} \label{eq: snapshot}
    \Xmat_K = 
    \begin{bmatrix}
        | & | &  & |\\
        \xvec_1 & \xvec_2 & \dots & \xvec_K\\
        | & | &  & |
    \end{bmatrix}
    \in \real^{N \times K}.
\end{equation}
Snapshot data may be obtained from a single simulation of the state at different times, or from multiple simulations from different initial conditions (or in parameterized systems, from different parameters). In \Cref{subsection: probform}, the snapshots will correspond to random initial conditions drawn from a given probability distribution. The \gls{svd} of $\Xmat_K$ is given by $\Xmat_K = \mathbf{\Phi} \mathbf{\Sigma} \mathbf{\Psi}^\top$, where the columns of $\mathbf{\Phi} \in \real^{N \times N}$ denote the left singular vectors of $\Xmat_K$, the columns of $\mathbf{\Psi} \in \real^{K \times K}$ denote the right singular vectors of $\Xmat_K$, and the diagonal entries $\sigma_1 \geq \sigma_2 \geq \dots$ in $\mathbf{\Sigma} \in \real^{N \times K}$ are the singular values of $\Xmat_K$ in non-increasing order. The POD basis of size $r$, denoted by $\mathbf{\Phi}_{r}$, is defined by the leading $r$ columns of $\mathbf{\Phi}$. We obtain a reduced state $\xhat$ by projecting onto the span of the \gls{pod} basis, i.e., $\hat{\xvec} = \mathbf{\Phi}_r^\top \xvec \in \real^r$. The POD state approximation $\xvec \approx \mathbf{\Phi}_{r}\hat{\xvec}$ is the best rank-$r$ approximation of the state~\cite{eckart1936approximation,mirsky1960symmetric}.

If we substitute the state approximation into the full model \eqref{eq: full_lq_model} and enforce the Galerkin orthogonality condition that the model residual be orthogonal to the approximation basis, i.e.,
\begin{equation*} 
    \mathbf{\Phi}^\top_{r}\big(\mathbf{\Phi}_{r}\dot{\hat{\xvec}}(t) - \Amat \mathbf{\Phi}_{r}\hat{\xvec}(t) - \Hmat (\mathbf{\Phi}_{r}\hat{\xvec}(t) \kron \mathbf{\Phi}_{r}\hat{\xvec}(t)) \big) = \mathbf{0},
\end{equation*}
we arrive at the following POD-Galerkin reduced model
\begin{equation} \label{eq: reduced_lq_model}
    \xhatdot(t) = \Ahat\xhat(t) + \Hhat (\xhat(t) \kron \xhat(t)), \qquad \xhat(0) = \xhat_0 = \mathbf{\Phi}^\top_r \xvec_0.
\end{equation}
The reduced linear operator is $\Ahat = \mathbf{\Phi}_{r}^\top \Amat \mathbf{\Phi}_{r}\in \real^{r \times r}$, the reduced quadratic operator is $\Hhat=\mathbf{\Phi}_{r}^\top\Hmat(\mathbf{\Phi}_{r}\kron\mathbf{\Phi}_{r})\in\real^{r \times r^2}$. We note that the reduced model \eqref{eq: reduced_lq_model} retains the linear-quadratic structure of the full order model \eqref{eq: full_lq_model}. The reduced order model \eqref{eq: reduced_lq_model} can be evaluated more efficiently since the size of the operators scales with reduced dimension $r$ rather than the full dimension $N$. 

We note that the POD-Galerkin model \eqref{eq: reduced_lq_model} is \textit{intrusive} and requires access to full-order operators $\Amat$ and $\Hmat$ to compute $\Ahat$ and $\Hhat$. However, in many settings, it is desirable to be able to compute reduced operators without intrusive access, e.g., when the underlying code is black-box, proprietary, or complex legacy code. In these settings, we consider access to $\Amat$ and $\Hmat$ to be non-trivial, and the reduced operators $\Ahat$ and $\Hhat$ must be obtained non-intrusively.

\subsection{Operator Inference} \label{subsection: opinf}
We now present the \gls{opinf} approach from~\cite{peherstorfer2016data}. The goal of \gls{opinf} is to fit the reduced operators $\Ahat$ and $\Hhat$ to reduced data in the least-squares sense. To obtain necessary reduced data, we require both the state data \eqref{eq: snapshot} as well as the paired state time derivative data
\begin{equation} \label{eq: snapshotderiv}
    \Xdot_K = 
    \begin{bmatrix}
        | & | &  & |\\
        \xdot_1 & \xdot_2 & \dots & \xdot_K\\
        | & | &  & |
    \end{bmatrix}
    \in \real^{N \times K} \quad \text{where} \quad
    \xdot_i = \Amat\xvec_i + \Hmat(\xvec_i \kron \xvec_i) \quad \text{where} \quad i = 1, \dots, K.
\end{equation}
The time derivative data can be derived directly from the numerical simulation or numerically approximated. We use the \gls{pod} basis $\mathbf{\Phi}_{r}$ to compute the reduced state and time derivative data as follows $\xhat_i = \mathbf{\Phi}_{r}^\top \xvec_i$ and $\xhatdot_i = \mathbf{\Phi}_{r}^\top \xdot_i$ for $i = 1, 2, \dots, K$ and define
\begin{equation} \label{eq: reduced snapshot}
    \Xhat_K = 
    \begin{bmatrix}
        | & | &  & |\\
        \xhat_1 & \xhat_2 & \dots & \xhat_K\\
        | & | &  & |
    \end{bmatrix}
    \in \real^{r \times K} \quad \text{and} \quad 
     \dot{\hat{\mathbf{X}}}_K = 
    \begin{bmatrix}
        | & | &  & |\\
        \dot{\hat{\mathbf{x}}}_1 & \dot{\hat{\mathbf{x}}}_2 & \dots & \dot{\hat{\mathbf{x}}}_K\\
        | & | &  & |
    \end{bmatrix}
    \in \real^{r \times K}
\end{equation}
We also define the quadratic state matrix to be $\hat{\Xmat}_{\kron}$ where 
\begin{equation} \label{eq: Xhatkron}
    \Xhat_K^{\kron} = 
    \begin{bmatrix}
        | & | &  & |\\
        \xhat_1 \kron \xhat_1 & \xhat_2 \kron \xhat_2& \dots & \xhat_K\kron \xhat_K\\
        | & | &  & |
    \end{bmatrix}
    \in \real^{r^2 \times K}.
\end{equation}

Such definitions allow us to set up the \gls{opinf} minimization problem as follows

\begin{equation} \label{eq:opinf}
\min_{\Ahat \in \real^{r\times r}, \Hhat\in \real^{r\times r^2}} \frac{1}{K}\sum_{i = 1}^{K} \norm{\xhatdot_i - \Ahat \!\xhat_i - \Hhat(\xhat_i \kron \xhat_i)}_2^2
=\min_{\Omat \in \real^{r\times (r + r^2)}} \frac{1}{K}\norm{\Dmat^\top_K \Omat - ~\Xhatdot_K^\top}^2_\text{F},
\end{equation}
where a data matrix of $K$ snapshots is $\Dmat_K = [\Xhat_K^\top\; (\Xhat_K^{\kron})^\top]^\top\in \real^{(r+r^2) \times K}$, the operator matrix is $\Omat~=~[\Ahat \; \Hhat]^\top ~\in~\real^{(r+r^2) \times r}$, and the norm $\norm{\cdot}_\text{F}$ denotes the Frobenius norm.

If we consider the rows of $\Xhatdot_K$, solving the least squares problem \eqref{eq:opinf} is equivalent to solving the following $r$ independent least-squares problems for each of the rows of the reduced operators or equivalently, each column of $\Omat$:
\begin{equation} \label{eq: decoupledopinf}
\min_{\ovec_i \in \real^{r + r^2}} \frac{1}{K}\sum_{i = 1}^K \norm{\Dmat^\top_K\ovec_i - \rvec^i_K}^2_2 \quad \text{for} \quad i = 1, \dots, r\quad \text{where} \quad \dot{\hat{\mathbf{X}}}^\top_K = 
    \begin{bmatrix}
        | & | &  & |\\
        \rvec^1_K & \rvec^2_K & \dots &\rvec^r_K\\
        | & | &  & |
    \end{bmatrix}
    \in \real^{K \times r}
\end{equation}
and $\ovec_i$ denotes the $i$-th column of $\Omat$. Therefore, we can obtain one column of the operator matrix $\Omat$ from the $r$ independent least squares problems with $(r+ r^2)$ degrees of freedom each.

We note that the redundancy in the Kronecker product means the minimizer of $\Hhat$ of \eqref{eq:opinf} is non-unique. When the structural redundancy in the Kronecker product is accounted for, the degrees of freedom in each least-squares problem can be reduced from $r + r^2$ to $r + \frac{r(r+1)}{2}$.

\subsection{Multifidelity Linear Regression} \label{subsection: mflr}
We now introduce the multifidelity linear regression approach from~\cite{qian2025multifidelity} which uses a control variate framework to reduce the variance in the estimation of unknown linear regression parameters. 

Let $\zvec \in \real^m$ denote a vector-valued input random variable with probability density function $\pi$. We denote by $y^{(1)} = f^{(1)}(\zvec)$ the scalar, \textit{high-fidelity} output values of interest, where $f^{(1)}: \real^m \to \real$ denotes the true, high-fidelity input-output relationship with cost $w_1$. Additionally, we denote by $y^{(2)} = f^{(2)}(\zvec)$ the scalar, \textit{low-fidelity} output values, where $f^{(2)}: \real^m \to \real$ denotes the low-fidelity input-output relationship with cost $w_2$ such that $w_2 < w_1$. We also define $\tilde{\uvec}: \real^m \rightarrow \real^d$ so that $\uvec = \tilde{\uvec}(\zvec)$ is a $d$-dimensional feature transformation of the input $\zvec$, a vector-valued random variable. The feature transformation $\uvec = \tilde{\uvec}(\zvec)$ forms the function basis, e.g. a polynomial basis, of our linear regression model. Linear regression aims to learn a model in the form
\begin{equation*}
    \hat{f}(\zvec, \boldsymbol{\boldsymbol{\beta}}) = \tilde{\uvec}(\zvec)^\top \boldsymbol{\beta},
\end{equation*}
by selecting $\boldsymbol{\beta} \in \real^d$ such that $\hat{f}(\zvec; \boldsymbol{\beta}) = \uvec^\top \boldsymbol{\beta} \approx y^{(1)}$. We introduce the following minimization problem to minimize the expected squared error over the entire probability distribution:
\begin{equation*}
    \boldsymbol{\beta}^* = \argmin_{\boldsymbol{\beta} \in \real^d} \E{\frac{1}{2}\norm{\hat{f}(\zvec;\boldsymbol{\beta}) - y^{(1)}}^2_2} = \argmin_{\boldsymbol{\beta} \in \real^d} \E{\frac{1}{2}\norm{\uvec^\top \boldsymbol{\beta} - y^{(1)}}^2_2},
\end{equation*}
where the exact minimizer is given by
\begin{equation} \label{eq: betastar}
    \boldsymbol{\beta}^* = \E{\uvec\uvec^\top}^{-1}\E{\uvec y^{(1)}}.
\end{equation}

In practice, computing exact expectations with respect to distribution $\pi$ is analytically intractable, so $\boldsymbol{\beta}^*$ from \eqref{eq: betastar} is approximated from data. The standard approach is to use a dataset $\{(\uvec_k, y^{(1)}_k)\}_{k = 1}^K$, where $y^{(1)}_k = f^{(1)}(\zvec_k)$ and $\uvec_k = \tilde{\uvec}(\zvec_k)$ for $\zvec_k$ drawn \gls{iid} from distribution $\pi$. However, in this work, we consider the setting where obtaining high-fidelity data $y_k^{(1)}$ for $k = 1, \dots, K$ is expensive for large sample sizes $K$, requiring repeated evaluation of the high-fidelity model $f^{(1)}$. In contrast, forming the features $\uvec_k$ does not require the expensive model $f^{(1)}$, and the cost to compute accurate estimates of $\E{\uvec\uvec^\top}^{-1}$ can be assumed to be negligible compared to the cost of estimating $\E{\uvec y^{(1)}}$. Therefore, we assume that exact values of $\E{\uvec\uvec^\top}^{-1}$ are available which allows the estimation of $\boldsymbol{\beta}^*$ to be unbiased.

The multifidelity approach assumes we have access to $K_1$ high-fidelity training points $\{(\uvec_k, y^{(1)}_k)\}_{k = 1}^{K_1}$, and $K_2$ low-fidelity training points $\{(\uvec_k, y^{(2)}_k)\}_{k = 1}^{K_2}$ where $K_2 > K_1$. The outputs $y^{(1)}_k$ and $y^{(2)}_k$ are the high and low-fidelity outputs evaluated from input $\zvec_k$, i.e., $y^{(1)}_k= f^{(1)}(\zvec_k)$ and $y^{(2)}_k= f^{(2)}(\zvec_k)$, and we note that the first $K_1$ evaluations of the low-fidelity model are paired with the $K_1$ evaluations of the high-fidelity model. Let $\Umat_{K_1}\in \real^{d \times K_1}$ and $\Umat_{K_2} \in \real^{d\times {K_2}}$ denote the matrix whose columns consist of the first $K_1$ and $K_2$ inputs respectively. Let $\yvec^{(1)}_{K_1} \in \real^{K_1}$ denote the vector whose elements are given by the first $K_1$ outputs of the high-fidelity model, and let $\yvec^{(2)}_{K_1} \in \real^{K_1}$ and $\yvec^{(2)}_{K_2} \in \real^{K_2}$ denote the vector whose elements are given by the first $K_1$ and $K_2$ outputs of the low-fidelity model respectively. We define the following multifidelity estimate of the exact minimizer \eqref{eq: betastar}:
\begin{equation} \label{eq: mfbeta}
    \hat{\boldsymbol{\beta}}^\text{MF} = \E{\uvec \uvec^\top}^{-1} \hat{\cvec}_{\uvec \yvec}^\text{MF} \quad \text{where} \quad \hat{\cvec}^\text{MF}_{\uvec \yvec} = \frac{1}{K_1}\Umat_{K_1}\yvec_{K_1}^\text{(1)} + \alpha \left(\frac{1}{K_2}\Umat_{K_2}\yvec_{K_2}^{(2)} -\frac{1}{K_1}\Umat_{K_1}\yvec_{K_1}^{(2)}\right), 
\end{equation}
where $\alpha \in \real$ is a control variate scalar coefficient. The multifidelity estimate \eqref{eq: mfbeta} defines the multifidelity estimator for the linear regression model:
\begin{equation*}
    \hat{f}(\zvec; \hat{\boldsymbol{\beta}}^\text{MF}) = \tilde{\uvec}(\zvec)^\top\hat{\boldsymbol{\beta}}^\text{MF}.
\end{equation*}

This multifidelity linear regression approach depends on three hyperparameters: the control variate scalar coefficient $\alpha$ and the number of high and low-fidelity samples $K_1$ and $K_2$ respectively.

To define the optimal choices for the control variate scalar coefficient $\alpha$, we note that the goal is to estimate $\E{\uvec y^{(1)}} = \E{\tilde{\uvec}(\zvec) f^{(1)}(\zvec)}$. Let $\gvec^{(1)}(\zvec) = \tilde{\uvec}(\zvec)f^{(1)}(\zvec) \in \real^d$ and $\gvec^{(2)}(\zvec) = \tilde{\uvec}(\zvec)f^{(2)}(\zvec) \in \real^d$, and define covariance matrices
\begin{equation} \label{eq:gams}
\mathbf{\Gamma}_{12} = \Cov{\gvec^{(1)}(\zvec), \gvec^{(2)}(\zvec)}, \quad \mathbf{\Gamma}_{22} = \Cov{\gvec^{(2)}(\zvec), \gvec^{(2)}(\zvec)}.
\end{equation}

Then, the optimal choice of $\alpha$ that minimizes the generalized variance $\Tr{\Cov{\hat{\cvec}^\text{MF}_{\uvec \yvec}, \hat{\cvec}^\text{MF}_{\uvec \yvec}}}$ where $\Tr{\cdot}$ denotes the matrix trace is
\begin{equation}
    \text{Optimal scalar choice:} \quad \alpha^* = \frac{\Tr{\mathbf{\Gamma}_{12}}}{\Tr{\mathbf{\Gamma}_{22}}}.
\end{equation}

Multifidelity linear regression as described in~\cite{qian2025multifidelity} offers heuristic choices for $K_1$ and $K_2$ and also considers optimal selection of matrix control variate coefficients. 

\section{Multifidelity Operator Inference} \label{section: mfopinf}
In this section, we recast the \gls{opinf} problem into the multifidelity linear regression setting in~\cref{subsection: probform}. Then, we introduce the \gls{mfopinf} framework for reduced operator estimation in \cref{subsection: mfopinf} and discuss selection of hyperparameters in \cref{subsection: hyperparameter} for \gls{mfopinf}.

\subsection{Problem Formulation} \label{subsection: probform}

Now, we will recast the \gls{opinf} problem in the multifidelity linear regression setting where we will consider data drawn from a probability distribution.

Let $\xvec_0 \in \real^{N}$ denote a random vector with known probability distribution $\pi_0$ and covariance matrix $\Cmat \in \real^{N \times N}$. The vector $\xvec_0$ represents the state of a dynamical system at time $t = 0$, and in our setting, $N$ is large and corresponds to a \textit{high-fidelity}, fine-mesh discretization. The \gls{pod} modes of the initial condition distribution correspond with the eigenvectors of covariance $\Cmat$. Thus, we form the columns of the reduced \gls{pod} basis $\mathbf{\Phi}_r$ with the leading $r$ eigenvectors $\{\unslant{\boldsymbol{\phi}}_i\}_{i = 1}^{r}$, i.e., 
\begin{equation*}
      \mathbf{\Phi}_r = 
    \begin{bmatrix}
        | & | &  & |\\
        \unslant{\boldsymbol{\phi}}_1 & \unslant{\boldsymbol{\phi}}_2 & \dots & \unslant{\boldsymbol{\phi}}_r\\
        | & | &  & |
    \end{bmatrix}.
\end{equation*}
In \Cref{subsection: ICs}, we provide a detailed discussion of obtaining the reduced \gls{pod} basis analytically from a \gls{grf}. 

Note that the ultimate goal of \gls{opinf} is to map the reduced state vector onto the reduced state time derivative. Let input random variable $\zvec$ be the \gls{pod} coefficients of the initial condition for the $N$-dimensional dynamical system, i.e., $\zvec = \mathbf{\Phi}_r^\top \xvec_0$. The high-fidelity output of interest $\yvec^{(1)} = \mathbf{f}^{(1)}(\zvec)$ is the vector-valued reduced time derivative evaluated at time $t = 0$, where $\mathbf{f}^{(1)}: \real^r \to \real^r$ denotes the true mapping from reduced state to reduced state time derivative with cost $w_1$. We observe that this is equivalent to simulating the $N$-dimensional model \eqref{eq: full_lq_model} and reducing the time derivative as follows:
\begin{equation} \label{eq: hf_model}
    \fvec^{(1)}(\zvec) = \mathbf{\Phi}^\top_r\xdot_0 = \mathbf{\Phi}^\top_r \big(\Amat \xvec_0 + \Hmat  (\xvec_0 \kron \xvec_0)\big) \quad \text{where} \quad \zvec = \mathbf{\Phi}_r^\top \xvec_0.
\end{equation}
In our setting, the time derivative $\xdot_0$ is numerically approximated by simulating the model \eqref{eq: full_lq_model} by one timestep with size $\Delta t$ and applying a forward time derivative approximation.

The low-fidelity model $\mathbf{f}^{(2)}: \real^r \to \real^r$ is a cheaper lower-fidelity approximation of the reduced time derivative with cost $w_2$ such that $w_2 <w_1$. In our setting, the low-fidelity model $\fvec^{(2)}$ is the result evaluating the full-order model \eqref{eq: full_lq_model} on a coarser grid $N_\ell < N$. We consider the following low-fidelity, $N_\ell$-dimensional system of \gls{ode}s:
\begin{equation} \label{eq: low_lq_model}
    \dot{\tilde{\xvec}}(t) = \tilde{\Amat} \tilde{\xvec}(t) + \tilde{\Hmat} (\tilde{\xvec}(t) \kron \tilde{\xvec}(t)), \qquad \tilde{\xvec}(0) = \tilde{\xvec}_0,
\end{equation}
where $\tilde{\Amat} \in \real^{N_\ell \times N_\ell}, \tilde{\Hmat}\in \real^{N_\ell \times N^2_\ell}$ are the low-fidelity linear and quadratic operators respectively. The low-fidelity initial condition $\tilde{\xvec}_0 \in \real^{N_\ell}$ is obtained by sub-sampling the high-fidelity initial condition $\xvec_0 \in \real^N$ on a coarser grid. The low-fidelity initial time derivative $\dot{\tilde{\xvec}}_0$ is numerically approximated by simulating the low-fidelity model \eqref{eq: low_lq_model} by one timestep with size $\Delta t_\ell$, where $\Delta t_\ell > \Delta t$, and applying a forward time derivative approximation. To match the dimension of the low-fidelity approximation to the high-fidelity grid, we define a linear interpolation operator $\Pmat \in \real^{N \times N_\ell}$ such that $\xdot_0 \approx \Pmat\,\dot{\tilde{\xvec}}_0$. Thus, the low-fidelity output of interest $\yvec^{(2)} = \mathbf{f}^{(2)}(\zvec)$ is the reduced time derivative approximation evaluated at time $t = 0$ and is equivalent to
\begin{equation} \label{eq: lf_model}
    \fvec^{(2)}(\zvec) = \mathbf{\Phi}^\top_r\Pmat \,\dot{\tilde{\xvec}}_0 = \mathbf{\Phi}^\top_r \Pmat \big(\tilde{\Amat} \tilde{\xvec}_0 + \tilde{\Hmat} (\tilde{\xvec}_0 \kron \tilde{\xvec}_0)\big).
\end{equation}

The \gls{opinf} feature transformation $\tilde{\dvec}: \real^r \to \real^{ r + r^2}$ is defined by constructing the linear and quadratic features in the reduced state as follows:
\begin{equation} \label{eq: opinf_feature}
    \tilde{\dvec}(\zvec) = \left[\zvec^\top \ (\zvec \kron \zvec)^\top \right]^\top \in \real^{r + r^2}.
\end{equation}

The goal of \gls{opinf} is to learn a model in the form \eqref{eq: reduced_lq_model} which can be rewritten using our defined feature transformation $\dvec = \tilde{\dvec}(\zvec)$ as follows:
\begin{equation*}
    \fhat(\zvec; \Omat) = \Omat^\top \tilde{\dvec}(\zvec) \quad \text{where} \quad\Omat = [\Ahat\; \Hhat]^\top \in \real^{(r + r^2) \times r}.
\end{equation*}
To find the operator matrix $\Omat$ that minimizes the squared-error over the entire probability distribution, the \gls{opinf} problem solves the following minimization:
\begin{equation*}
    \Omat^* = \argmin_{\Omat \in \real^{(r + r^2)\times r }} \E{\frac{1}{2}\norm{\mathbf{\fhat}(\zvec;\Omat) - \yvec^{(1)} }^2_2} = \argmin_{\Omat \in \real^{(r + r^2)\times r }} \E{\frac{1}{2}\norm{\dvec^\top \Omat - (\yvec^{(1)})^\top}^2_2},
\end{equation*}
in which the exact minimizer is given by
\begin{equation} \label{eq: exacto}
    \Omat^* = \E{\dvec \dvec^\top}^{-1} \E{\dvec (\yvec^{(1)})^\top}.
\end{equation}

The standard approach for approximating \eqref{eq: exacto} is to use a dataset comprised of $K$ paired snapshots $\{(\dvec_k, \yvec_k^{(1)})\}_{k =1}^K$, where $ \yvec^{(1)}_k = \fvec^{(1)}(\zvec_k)$ and $\dvec_k =\tilde{\dvec}(\zvec_k)$ for $\zvec_k$ drawn \gls{iid} from distribution $\pi_0$. However, as discussed in~\Cref{subsection: mflr}, high-fidelity snapshots $\yvec^{(1)}_k$ from $\fvec^{(1)}$ are expensive to obtain when the sample size $K$ is large due to repeated evaluation of an expensive model. In contrast, computing features $\dvec_k = \tilde{\dvec}(\zvec_k)$ do not require evaluation of $\fvec^{(1)}$ and operate only on reduced state representations. Thus, features $\dvec_k$ may be cheaply formed in comparison, and we can assume exact values of $\E{\dvec \dvec^\top}^{-1}$ are available. 

Let $\Dmat_K \in \real^{(r + r^2) \times K}$ denote the matrix whose columns consist of $K$ input features, and let $\Ymat^{(1)}_K$ denote the matrix whose columns consist of the paired $K$ output snapshots as follows:
\begin{equation} \label{eq: DY def}
    \Dmat_K = 
    \begin{bmatrix}
        | & | &  & |\\
        \dvec_1 & \dvec_2 & \dots & \dvec_K\\
        | & | &  & |
    \end{bmatrix}
    \in \real^{(r+r^2) \times K} \quad \text{and} \quad 
    \Ymat^{(1)}_K = 
    \begin{bmatrix}
        | & | &  & |\\
        \yvec^{(1)}_1 & \yvec^{(1)}_2 & \dots & \yvec^{(1)}_K\\
        | & | &  & |
    \end{bmatrix}
    \in \real^{r \times K}
\end{equation}
We note that the above definitions are consistent with those in \eqref{eq:opinf} within \cref{subsection: opinf}, with the new interpretation of $\Xhatdot_K$ as high-fidelity output matrix $\Ymat^{(1)}_K$ and using snapshot data from randomly drawn initial conditions.

Using matrices defined in \eqref{eq: DY def}, we obtain a single, high-fidelity (HF) estimator for $\Omat^*$ from $K$ snapshots:
\begin{equation*} 
    \Ohat^\text{HF} = \E{\dvec \dvec^\top}^{-1} \Chat^\text{HF}_{\dvec \yvec} \quad \text{where} \quad \Chat^\text{HF}_{\dvec \yvec} = \frac{1}{K}\Dmat_K (\Ymat^{(1)}_K)^\top
\end{equation*}

Since the \gls{opinf} problem decouples into $r$ independent linear least squares problems, we can alternatively approximate each column of $\Omat^*$ as follows:
\begin{equation} \label{eq: hfopinf_decoupled}
    \ohat_i^\text{HF} = \E{\dvec \dvec^\top}^{-1} \chat^\text{HF}_{\dvec \yvec, i} \quad \text{where} \quad \chat^\text{HF}_{\dvec \yvec, i} = \frac{1}{K}\Dmat_K \rvec^{(1), i}_K,
\end{equation}
where $\rvec^{(1), i}_K \in \real^{K \times 1}$ denotes the $i$-th row of $\Ymat^{(1)}_K$, i.e.,
\begin{equation*}
    (\Ymat^{(1)}_K)^\top = 
    \begin{bmatrix}
        | & | &  & |\\
        \rvec^{(1), 1}_K & \rvec^{(1), 2}_K & \dots &\rvec^{(1), r}_K\\
        | & | &  & |
    \end{bmatrix}
    \in \real^{K \times r}
\end{equation*}
which results in the single, high-fidelity \gls{opinf} model:
\begin{equation} \label{eq: sfopinf}
    \fhat(\zvec; \Ohat^\text{HF}) = (\Ohat^\text{HF})^\top \tilde{\dvec}(\zvec).
\end{equation}
The estimator \eqref{eq: hfopinf_decoupled} is subject to high generalized variance $\Tr{\Cov{\ohat_i^\text{HF}, \ohat_i^\text{HF}}}$ when data is scarce, i.e., sample size $K$ is low. In practice, this corrupts the model with high variance, and its predictions in \eqref{eq: sfopinf} are likely to drift far from the ideal model $\fhat(\zvec; \Omat^*)$. As a result, the goal of \gls{mfopinf} is to use ample low-fidelity data from $\fvec^{(2)}$ to form an estimator of \eqref{eq: exacto} with a lower variance than \eqref{eq: hfopinf_decoupled}. 

\subsection{Multifidelity Operator Inference Method} \label{subsection: mfopinf}
Since the \gls{opinf} problem decouples into $r$ independent linear least squares problems \eqref{eq: decoupledopinf}, our strategy is to apply multifidelity linear regression from \cref{subsection: mflr} to each of those least squares problems. 

Denote $f^{(1)}_i: \real^{r} \to \real$ and $f^{(2)}_i:\real^{r} \to \real$ as the $i$-th element in the high and low-fidelity model respectively. The multifidelity linear regression approach to \gls{opinf} assumes we have access to $K_{1, i}$ input features and high-fidelity training outputs $\{(\dvec_k, r^{(1), i}_k)\}_{k = 1}^{K_{1, i}}$, and $K_{2, i}$ low-fidelity training outputs $\{(\dvec_k,  r^{(2), i}_k)\}_{k = 1}^{K_{2, i}}$ where low-fidelity data is more available, i.e., $K_{2, i} > K_{1, i}$. The outputs $r^{(1), i}_k$ and $r^{(2), i}_k$ are the $i$-th element of the high and low-fidelity snapshot evaluated from input $\zvec_k$ respectively, i.e., $r^{(1), i}_k = f^{(1)}_i(\zvec_k)$ and $r^{(2), i}_k = f^{(2)}_i(\zvec_k)$. We note that the first $K_{1, i}$ evaluations of the low-fidelity model are paired with the $K_{1, i}$ evaluations of the high-fidelity model. Let $\Dmat_{K_{1, i}}\in \real^{(r + r^2) \times K_{1, i}}$ and $\Dmat_{K_{2, i}}\in \real^{(r + r^2) \times K_{2, i}}$ denote the feature matrix whose columns consist of the first $K_{1, i}$ and $K_{2, i}$ input feature transformations respectively. Let $\rvec^{(1), i}_{K_{1, i}} \in \real^{K_{1, i}}$ denote the vector whose elements are given by the first $K_{1, i}$ outputs from the $i$-th element of the high-fidelity snapshots, and let $\rvec^{(2), i}_{K_{1, i}} \in \real^{K_{1,i}}$ and $\rvec^{(2)}_{K_{2, i}} \in \real^{K_{2, i}}$ denote the vector whose elements are given by the first $K_{1, i}$ and $K_{2, i}$ outputs from the $i$-th element of the low-fidelity snapshots respectively. Then, for $i = 1, \dots, r$, we define the following \gls{mfopinf} estimator:
\begin{equation} \label{eq: mfopinf_decoupled}
    \ohat_i^\text{MF} = \E{\dvec \dvec^\top}^{-1} \chat^\text{MF}_{\dvec \yvec, i} \quad \text{where} \quad \chat^\text{MF}_{\dvec \yvec, i} = \frac{1}{K_{1, i}}\Dmat_{K_{1, i}} \rvec^{(1), i}_{K_{1, i}} + \alpha_i\left( \frac{1}{K_{2, i}}\Dmat_{K_{2, i}} \rvec^{(2), i}_{K_{2, i}} - \frac{1}{K_{1, i}}\Dmat_{K_{1, i}} \rvec^{(2), i}_{K_{1, i}}\right), 
\end{equation}
where $\alpha_i$ denotes the control variate scalar coefficient. Like the multifidelity linear regression approach (\Cref{subsection: mflr}), the estimator is unbiased, i.e., $\E{\Ohat^\text{MF}} =\E{\Omat^*}$.We can construct the multifidelity-estimated $\Ohat^\text{MF}$ and extract the associated linear and quadratic reduced operators as follows:
\begin{equation*}
    \Ohat^\text{MF} = 
    \begin{bmatrix}
        | & | &  & |\\
        \ohat^\text{MF}_1 & \ohat^\text{MF}_2 & \dots & \ohat^\text{MF}_r\\
        | & | &  & |
    \end{bmatrix} = \begin{bmatrix}
        (\Ahat^\text{MF})^\top\\
        (\Hhat^\text{MF})^\top
    \end{bmatrix} \in \real^{(r + r^2) \times r}.
\end{equation*}

The multifidelity estimate \eqref{eq: mfopinf_decoupled} defines the \gls{mfopinf} model:
\begin{equation*}
    \fhat(\zvec; \Ohat^\text{MF}) = (\Ohat^\text{MF})^\top \tilde{\dvec}(\zvec) \quad \text{or equivalently} \quad \xhatdot = \Ahat^\text{MF}\xhat + \Hhat^\text{MF} (\xhat \kron \xhat).
\end{equation*}

\subsection{Optimal Control Variate Coefficients and Sample Allocations} \label{subsection: hyperparameter}
We note that the framework outlined in \cref{subsection: mfopinf} requires the user to choose hyperparameters $\alpha_i$, $K_{1, i}$, and $K_{2, i}$ for estimation of $\ohat^\text{MF}_i$. 
For the control variate coefficient $\alpha_i$, an optimal choice which minimizes the generalized variance $\Tr{\Cov{\ohat^\text{MF}_i, \ohat^\text{MF}_i}}$ is given by~\cite{qian2025multifidelity}
\begin{equation} \label{eq: optimalalpha}
    \alpha_i^* = \frac{\Tr{\mathbf{\Gamma}^i_{12}}}{\Tr{\mathbf{\Gamma}^i_{22}}},
\end{equation}
where for $i = 1, \dots, r$, we define $\gvec_i^{(1)}(\zvec) = \tilde{\dvec}(\zvec)f_i^{(1)}(\zvec) \in \real^{r + r^2}$ and $\gvec_i^{(2)}(\zvec) = \tilde{\dvec}(\zvec)f_i^{(2)}(\zvec) \in \real^{r + r^2}$, and the associated covariance matrices
\begin{equation*}
    \mathbf{\Gamma}^i_{11} = \Cov{\gvec_i^{(1)}(\zvec), \gvec_i^{(1)}(\zvec)}, \quad \mathbf{\Gamma}^i_{12} = \Cov{\gvec_i^{(1)}(\zvec), \gvec_i^{(2)}(\zvec)}, \quad \text{and} \quad \mathbf{\Gamma}^i_{22} = \Cov{\gvec_i^{(2)}(\zvec), \gvec_i^{(2)}(\zvec)}.
\end{equation*}

To determine the sample allocation strategy, the work~\cite{rex2026analyticallyoptimalsampleallocations} provides expressions for $K_{1,i}$ and $K_{2,i}$ minimizing the generalized variance in terms of the following coefficients:
\begin{equation*}
    b_{1, i} = \Tr{\E{\dvec \dvec^\top}^{-1}\left(\mathbf{\Gamma}^i_{11} - \mathbf{\Gamma}^i_{12}(\mathbf{\Gamma}^i_{22})^{-1}(\mathbf{\Gamma}^i_{12} )^\top\right)}, \quad b_{2, i} = \Tr{\E{\dvec \dvec^\top}^{-1} \mathbf{\Gamma}^i_{12}(\mathbf{\Gamma}^i_{22})^{-1}(\mathbf{\Gamma}^i_{12})^\top}.
\end{equation*}
Then, let $p$ be the computational budget, and recall that the associated computational cost of evaluating $\fvec^{(1)}$ and $\fvec^{(2)}$ are $w_1$ and $w_2$ respectively. The optimal choices for $K_{1, i}$ and $K_{2, i}$ that minimize the generalized variance of the column-wise estimator $\Tr{\Cov{\ohat^\text{MF}_i, \ohat^\text{MF}_i}}$ are~\cite{rex2026analyticallyoptimalsampleallocations}
\begin{equation} \label{eq: sample allocation}
    K^*_{1, i} = \max \left( 1, \left\lfloor \frac{p\sqrt{b_{1, i}/w_1}}{\sqrt{b_{1, i} w_1} + \sqrt{b_{2, i}w_2}}\right\rfloor\right), \quad 
    K^*_{2, i} = \max \left( K^*_{1, i}+1 , \left\lfloor \frac{p - K^*_{1, i}w_1}{w_2}\right\rfloor \right).
\end{equation}
We note that these choices are optimal for minimizing the variance of $\Tr{\Cov{\ohat^\text{MF}_i,\ohat^\text{MF}_i}}$ \textit{for each} $i$ and thus vary over $i = 1,2,\ldots,r$ (corresponding to the $i$th row of the reduced operators). However, since every simulation of the high- or low-fidelity model generates samples for estimation of all $r$ rows of the reduced operators, it does not make sense to use a different sample allocation for each row. 
In order to determine an overall sample allocation strategy that stays within budget, we use sample allocation from the row requiring the maximum number of high-fidelity samples: 
\begin{equation} \label{eq: sampleallocation}
    K_1 = K_{1, i^*} \; \text{and} \; K_2 = K_{2, i^*} \; \text{where} \; i^* = \argmax_{i \in \{1, \dots, r\}} K_{1, i}.
\end{equation}
We then use this choice of $K_1,K_2$ for the multifidelity estimation of every row of the reduced operators, as follows:
\begin{equation} \label{eq: mfopinf_coupled}
    \Ohat^\text{MF} = \E{\dvec \dvec^\top}^{-1} \Chat^\text{MF}_{\dvec \yvec} \quad \text{where} \quad \Chat^\text{MF}_{\dvec \yvec} = \frac{1}{K_{1}}\Dmat_{K_{1}} (\Ymat^{(1)}_{K_1})^\top + \left( \frac{1}{K_{2}}\Dmat_{K_{2}} (\Ymat^{(2)}_{K_2})^\top - \frac{1}{K_{1}}\Dmat_{K_{1}} (\Ymat^{(2)}_{K_1})^\top\right) \Amat_\alpha, 
\end{equation}
where $\Amat_\alpha = \text{diag}(\alpha_1, \dots, \alpha_r)$ is the control variate coefficient for the low-fidelity term, right-multiplied to account for column-wise estimation.

\section{Numerical Experiments} \label{section: numerics}
In this section, we discuss the distribution of initial conditions drawn in \cref{subsection: ICs} for numerical experiments of multifidelity operator inference on two synthetic \gls{pde} examples: the one-dimensional heat equation in~\cref{subsection: heateq} and the one-dimensional viscous Burgers' equation in \cref{subsection: burgeq}. We also present results on the \gls{kse} in \cref{subsection: KSE}.

\subsection{Distribution of initial conditions} \label{subsection: ICs}

In our numerical experiments, we consider a one-dimensional parabolic \gls{pde} with periodic boundary conditions on the spatial domain \( \Omega = [0, L] \subset \mathbb{R} \), with spatial coordinate \( \eta \in \Omega \), and parameter \( \mu \),
\begin{equation}\label{eq:general-pde}
    \frac{\partial}{\partial t} x(\eta, t; \mu)
    = g\!\left(x,\, \frac{\partial x}{\partial \eta},\, \frac{\partial^2 x}{\partial \eta^2},\, \eta,\, t;\, \mu \right),
    \qquad
    \begin{cases}
        x(0, t; \mu) = x(L, t; \mu), & \forall t \ge 0, \\
        x(\eta, 0; \mu) = x_0(\eta; \mu), & \forall \eta \in \Omega,
    \end{cases}
\end{equation}
where the function \(g\) determines the dynamics of the system. Discretizing such a \gls{pde} is one way the high-dimensional system~\eqref{eq: full_lq_model} arises.

The initial condition \( x_0(\eta) \) is modeled as a \gls{grf}. By definition, a random field \( x_0(\eta) \) is Gaussian if and only if, for any finite set of points \( \{\eta_1,\ldots,\eta_N\} \subset \Omega \), the vector \( \mathbf{x}_0 = \bigl[x_0(\eta_1), \ldots, x_0(\eta_N)\bigr]^\top \in \mathbb{R}^N \) is multivariate Gaussian~\cite{williams2006gaussian}.

In this work, we consider a zero-mean \gls{grf} with covariance operator \( \mathcal{C} \), and we generate samples using a truncated \gls{kl} expansion~\cite{karhunen1947lineare,loeve1978sums}. This expansion represents the random field as a finite sum of orthogonal spatial modes weighted by independent Gaussian random variables. Under periodic boundary conditions, the \gls{kl} expansion takes the form
\begin{equation}\label{eq:kl-expansion}
    x_0(\eta)
    = \sum_{k=1}^{N/2}  \xi_{2k}\,  \lambda_{2k}\, \phi_{2k}(\eta)
    + \xi_{2k-1}\, \lambda_{2k-1} \, \phi_{2k-1}(\eta),
\end{equation}
where \( \{\phi_k\} \) and \( \{\lambda_k\} \) are the eigenfunctions and eigenvalues of the covariance operator \( \mathcal{C} \), and \( \xi_k \sim \mathcal{N}(0,1) \), i.i.d.

For the class of covariance operators considered here, the eigenpairs are given explicitly for \( k = 1, \ldots, N/2 \) by
\begin{equation}\label{eq:grf-eigenpair}
    \phi_{2k}(\eta) = \sqrt{\frac{2}{L}} \cos\!\left( \frac{2\pi k \eta}{L} \right),
    \quad
    \phi_{2k-1}(\eta) = \sqrt{\frac{2}{L}} \sin\!\left( \frac{2\pi k \eta}{L} \right),
    \quad
    \lambda_{2k}^2 = \lambda_{2k-1}^2
    = \tau^{2s - 1}
    \left( \frac{4\pi^2 k^2}{L^2} + \tau^2 \right)^{-s}.
\end{equation}
Note that the hyperparameters \( \tau > 0 \) and \( s > 1/2 \) control the statistical properties of the \gls{grf}. The parameter \( \tau \) determines the correlation length (with characteristic length scale \( 1/\tau \)), such that larger values of \( \tau \) yield more rapidly varying fields. The parameter \( s \) controls the regularity (smoothness) of the field: larger values of \( s \) result in faster eigenvalue decay and therefore smoother realizations.

Given a set of points \( \{\eta_1,\ldots,\eta_N\} \) corresponding to the spatial discretization, a realization of the truncated \gls{grf} is represented by the vector
\begin{equation*}
\mathbf{x}_0 = \bigl[x_0(\eta_1), \ldots, x_0(\eta_N)\bigr]^\top \in \mathbb{R}^N.
\end{equation*}
Since the \gls{kl} expansion~\eqref{eq:kl-expansion} diagonalizes the covariance operator, the eigenfunctions \( \{\phi_k\} \) coincide with the \gls{pod} modes of the \gls{grf}. Hence, the reduced \gls{pod} basis \( \boldsymbol{\Phi}_r \in \mathbb{R}^{N \times r} \) formed by the first \( r \) modes yields \( \mathbf{z} = \boldsymbol{\Phi}_r^\top \mathbf{x}_0 \sim \mathcal{N}(\mathbf{0}, \boldsymbol{\Lambda}_r) \),
where \( \boldsymbol{\Lambda}_r = \mathrm{diag}(\lambda^2_1, \ldots, \lambda^2_r) \) contains the \( r \) largest eigenvalues of~\eqref{eq:grf-eigenpair}.

\begin{figure}
    \centering
    \includegraphics[width=0.6\linewidth]{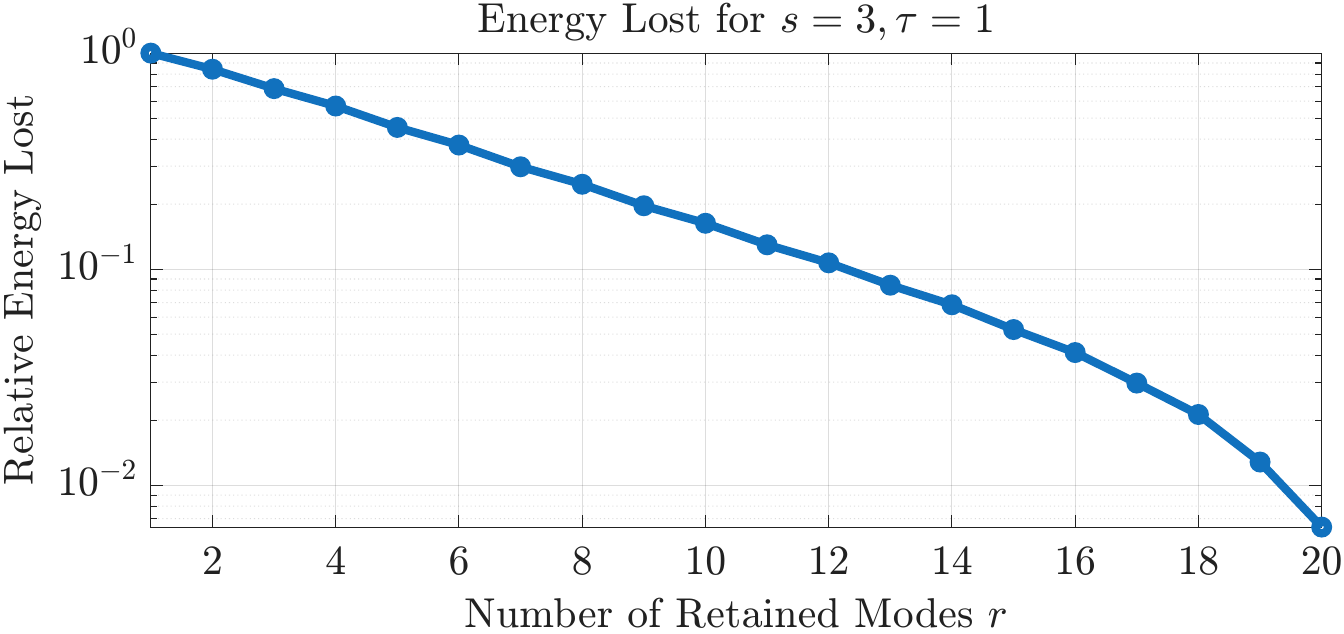}
    \caption{Relative energy loss from truncating to $r$ modes. }
    \label{fig:energy_loss}
\end{figure}

We can use the analytically obtained \gls{pod} and eigenvalues to calculate the relative energy loss when truncating to $r$ modes:
\begin{equation}
    \text{Relative Energy Lost:} \quad \frac{\sum_{i = r + 1}^N\lambda^2_i}{\sum_{i =1}^N\lambda^2_i}.
\end{equation}
\Cref{fig:energy_loss} demonstrates the energy loss for choosing hyperparameters $s = 3$ and $\tau = 1$ in \eqref{eq:grf-eigenpair}.

\subsection{1D Heat Equation} \label{subsection: heateq}


In this numerical experiment, we consider the heat equation, a diffusion model that describes how a quantity of interest diffuses through a domain, i.e., heat through a rod~\cite{cannon_one-dimensional_1984}. The 1D heat equation with periodic boundary conditions has form \eqref{eq:general-pde} with 
\begin{equation} \label{eq: heatPDE}
    g\left(\frac{\partial^2 x}{\partial \eta^2};\, \mu\right) = \mu \frac{\partial^2 }{\partial \eta^2}x(\eta, t)
\end{equation}
where $x(\eta, t)$ is the heat at a given point, $t$ is the time, $\eta$ is the spatial coordinate, and $\mu$ is the thermal diffusivity coefficient. Equation \eqref{eq: heatPDE} can be derived from Fourier's law for an isotropic (uniform) medium.

\subsubsection{Setup}
We define the domain and parameters of Equation \eqref{eq:general-pde} for \eqref{eq: heatPDE} as follows. Let $\Omega = [0, L] \subset \real$ where $L = 1$ be the spatial domain and $t \geq 0$ the time domain. We set the parameter to $\mu = 0.01$. Discretization of the spatial domain $\Omega$ of \eqref{eq: heatPDE} with $N=128$ uniformly-spaced points using the finite difference method yields the following system of \gls{ode}s:
\begin{equation} \label{eq: linearODE}
    \xdot(t) = \Amat \xvec(t)
\end{equation}
with state vector $\xvec(t) \in \real^N$ and linear operator $\Amat \in \real^{N \times N}$.

We consider the initial condition described in \eqref{eq:kl-expansion}, and let eigenvalue parameters be $\tau = 1 \;\text{and}\; s = 3$. We test reduced dimensions $r = 1, \dots, 12$ in our experiments which corresponds to retaining about $90$\% of the energy according to \Cref{fig:energy_loss}. To collect training data, we draw coefficients $\{\xi_i\}_{i = 1}^{N}$ \gls{iid} from $\mathcal{N}(0, 1)$ for each initial condition. 

\begin{figure}[h]
    \centering%
    \includegraphics[width=\linewidth]{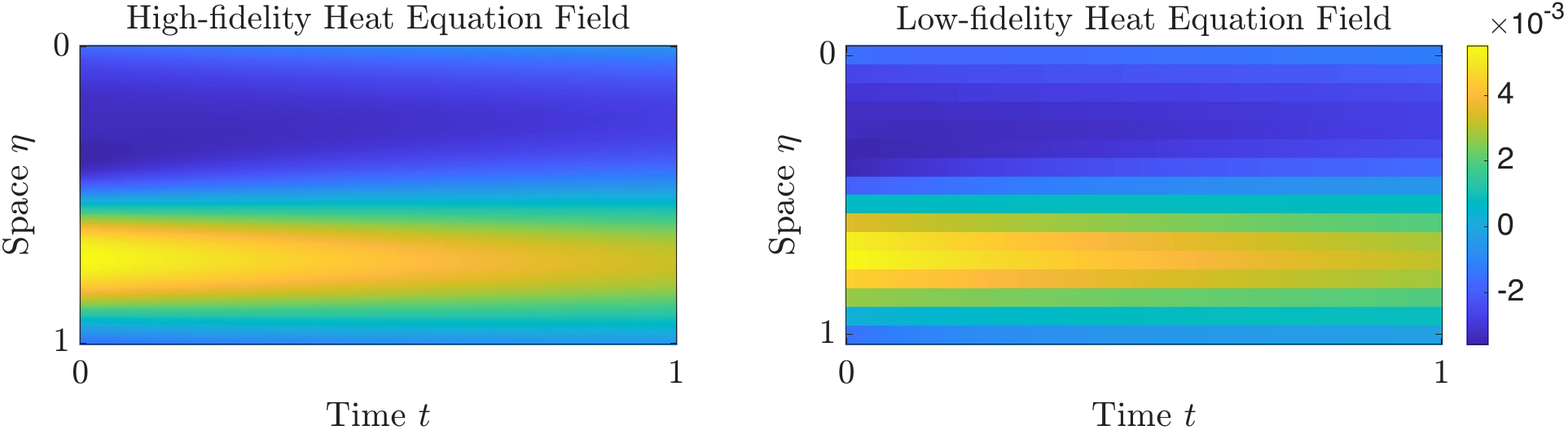}
    \caption{Visualization of high versus low-fidelity discretization of the 1D Heat Equation finite difference model, integrated to final time $t = 1$ with $\Delta t = 10^{-3}$. A random initial condition is drawn from training data set as described in the experimental setup.}
    \label{fig: heat_fields}
\end{figure}

High-fidelity time derivative data $\{\xdot_k\}$ arises from simulating \eqref{eq: linearODE} for the initial condition. The low-fidelity data arises from discretizing the spatial domain $\Omega$ of \eqref{eq: heatPDE} into $N_\ell = 16$ uniformly-spaced points using the finite difference method, resulting into the following system of \gls{ode}s:
\begin{equation} \label{eq: low_linearODE}
    \dot{\tilde{\xvec}}(t) = \tilde{\Amat} \tilde{\xvec}(t) 
\end{equation}
with state vector $\tilde{\xvec}(t) \in \real^{N_\ell}$ and linear operator $\tilde{\Amat}\in \real^{N_\ell \times N_\ell}$. The low-fidelity data $\{\dot{\tilde{\xvec}}_k\}$ is collected by simulating \eqref{eq: low_linearODE}. A visualization of the high- and low-fidelity discretization of a solution field integrated using the backwards Euler scheme to a final time of $t = 1$ with $\Delta t = 10^{-3}$ is provided in \Cref{fig: heat_fields}.

The POD basis $\mathbf{\Phi}_r$ is obtained analytically from the \gls{grf} covariance, and we reduce high and low-fidelity data to obtain $\{\yvec^{(1)}_k\}$ and $\{\yvec^{(2)}_k\}$ respectively as demonstrated in \eqref{eq: hf_model} and \eqref{eq: lf_model}. In this linear setting, the high and low-fidelity quadratic operators are zero, i.e., $\Hmat, \tilde{\Hmat} = \mathbf{0}$, and the feature transformation only contains the identity feature $\dvec_k =\zvec_k$. Model statistics for hyperparameter selection as discussed in \cref{subsection: hyperparameter} are calculated assuming perfect statistics are available. Statistics are obtained by generating a large dataset of $3 \times 10^6$ snapshots.

The standard \gls{opinf} \eqref{eq: hfopinf_decoupled} and \gls{mfopinf} \eqref{eq: mfopinf_decoupled} estimators for $\ohat_i^*$ are computed by drawing \gls{iid} snapshots from the \gls{grf} within budget constraint $p$. We normalize the model costs so that the cost to evaluate one high-fidelity snapshot has cost $w_1 = 1.00$ and the cost to evaluate one low-fidelity snapshot has cost $w_2 = 1.56 \times 10^{-2}$, where costs are calculated from \gls{flops} needed to evaluate a snapshot.

We train $M_\text{rep} = 100$ replicates of \gls{opinf} operators and \gls{mfopinf} operators, where replicates are trained with different realizations of random training data. To determine the sample size $K$ for training single, high-fidelity \gls{opinf} operators, we divided the budget by the high-fidelity cost $K = p/w_1$. Sample allocation for \gls{mfopinf} $K_{1}, K_{2}$ is determined using \eqref{eq: sampleallocation}, and the scaling coefficient $\{\alpha_i\}_{i = 1}^r$ is determined using \eqref{eq: optimalalpha}. 

We generate testing data with $M_\text{test} = 20$ \gls{iid} drawn initial conditions from the training region as described in \eqref{eq:kl-expansion}. The system of \gls{ode}s \eqref{eq: linearODE} is marched forward in time for each new initial condition with the implicit Euler scheme at timestep size $\Delta t = 10^{-3}$ with final time $t = 1$ to generate a test dataset of $M_\text{test}$ heat fields $\{\Xmat_i\}_{i=1}^{M_\text{test}}$. We use the learned models to reconstruct the additional data by simulating them from the same new initial conditions $\{\mathbf{\Phi}_r \Xhat_{j, i}\}_{i=1}^{M_\text{reps}}$, where $\Xhat_{j, i}$ is the predicted reduced-dimensional heat field from the $j^\text{th}$ operator replicate and $i^\text{th}$ test initial condition. We report the averaged state error of the reduced model reconstruction defined as:
\begin{equation} \label{eq: state_error}
    \text{Mean Relative State Error: } 
    \frac{1}{M_\text{test}} \frac{1}{M_\text{reps}}\sum_{i = 1}^{M_\text{test}}\sum_{j = 1}^{M_\text{reps}}\frac{\norm{(\Xmat_i-\mathbf{\Phi}_r \Xhat_{j, i})}^2_\text{F}}{\norm{\Xmat_i}^2_\text{F}}.
\end{equation}
which obtains the relative integrated error over the spatial and temporal domain. The averaged state error is calculated for $\Xhat_{j, i}$ obtained from both single, high-fidelity trained \gls{opinf} models and \gls{mfopinf} trained models, for reduced dimension sizes $r = 1, \dots, 10$, and for budget sizes $p = 10,  100$, $1000$, $10000$.

\subsubsection{Results}
\begin{figure}[t]
    \centering
    \includegraphics[width=\linewidth]{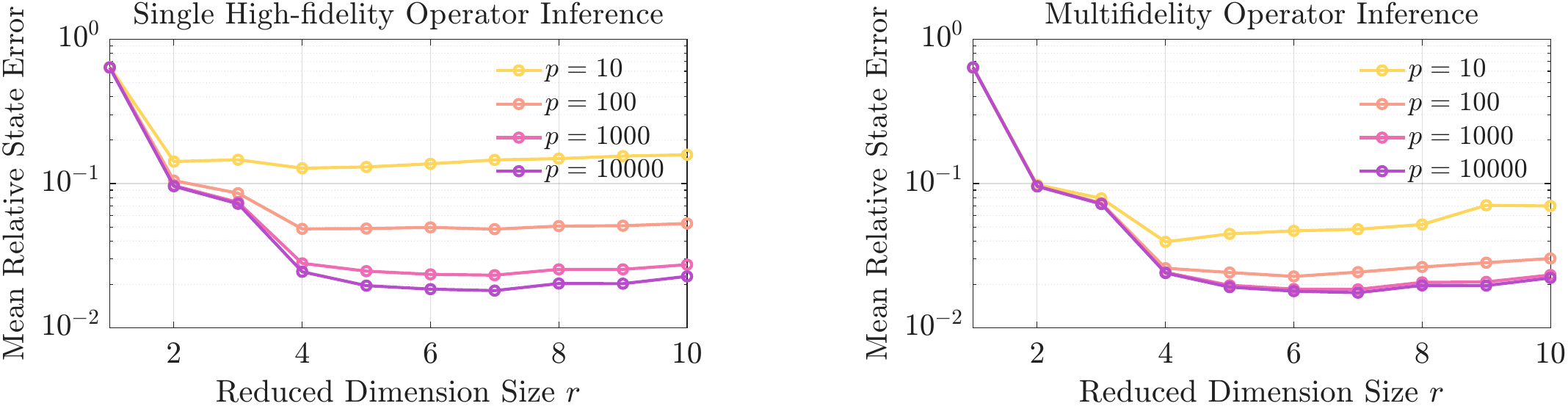}
    \caption{Comparison of mean relative state error between the standard single, high-fidelity \gls{opinf} (left) and the new \gls{mfopinf} field predictions for the 1D Heat Equation example.}
    \label{fig: heatstateerror}
\end{figure}

\Cref{fig: heatstateerror} shows the average relative state error over trained model replicates and over different test fields for different training budgets and reduced dimension size for single, high-fidelity \gls{opinf} (left) and \gls{mfopinf} (right). Across all reduced dimensions $r = 1, \dots, 10$, we observe a lower mean relative state error for the \gls{mfopinf} results compared to the single, high-fidelity \gls{opinf}. The difference in performance is most prominent in the lower budget sizes where high-fidelity data is scarce, which demonstrates how low-fidelity data can compensate for the lack of high-fidelity data.

\begin{figure}[t]
    \centering
    \includegraphics[width=0.7\linewidth]{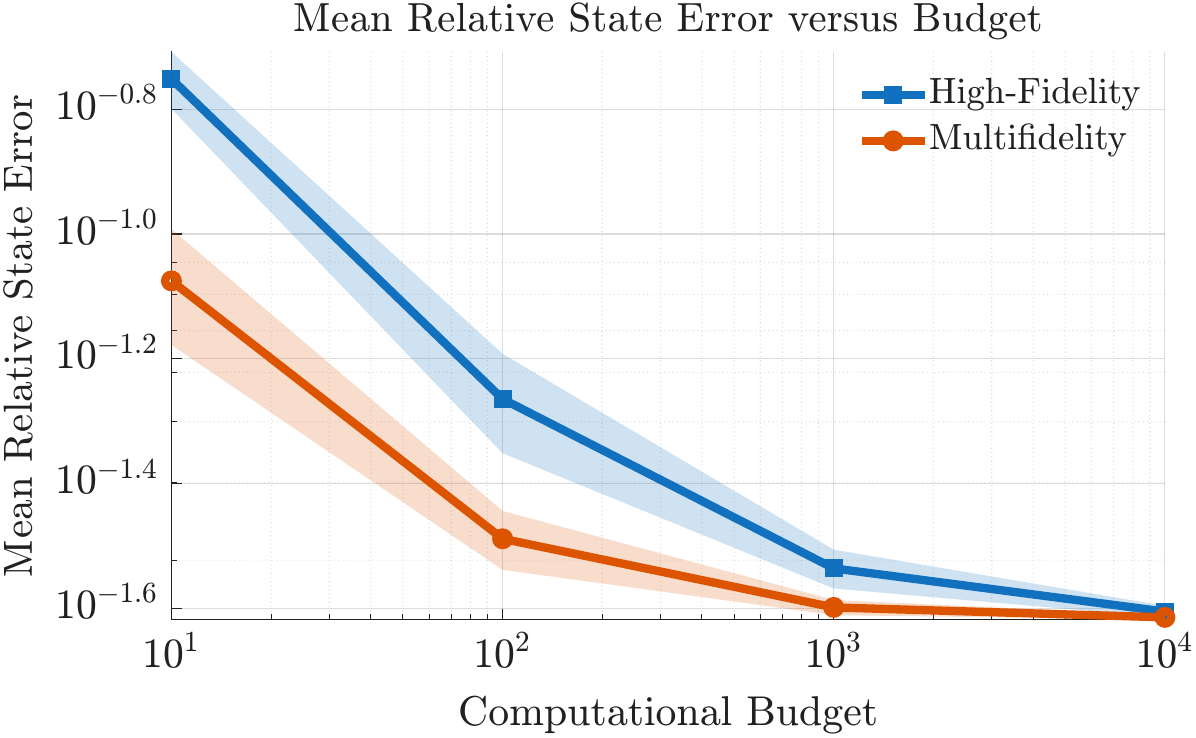}
    \caption{Comparison of mean relative state error $\pm$ one standard deviation of the operator replicates between the standard single, high-fidelity \gls{opinf} and \gls{mfopinf} learned with reduced dimension size $r = 10$ for the 1D Heat Equation example.}
    \label{fig: heatstateerror_12}
\end{figure}

In \Cref{fig: heatstateerror_12}, we plot the mean relative state error for just reduced dimension size $r = 10$ with respect to budget sizes $p = 10, 100, 1000, 10000$. For low budgets, we observe the greatest benefit in implementing \gls{mfopinf} over the standard \gls{opinf}, again demonstrating the efficacy of utilizing a multifidelity approach when high-fidelity data is scarce.

We note that in both \Cref{fig: heatstateerror} and \Cref{fig: heatstateerror_12} the mean relative state errors are higher orders of magnitude than standard model reduction/\gls{opinf} literature. This can be attributed to our chosen distribution of initial conditions as well as extrapolating past training data by $1000$ time steps.

\subsection{1D Viscous Burgers' Equation} \label{subsection: burgeq}
In this numerical example, we consider the viscous Burgers' equation, a simplified fluid dynamics model that describes the motion of a one-dimensional viscous fluid~\cite{mises_mathematical_1948}. The one-dimensional viscous Burgers' equation with periodic boundary conditions has the form \eqref{eq:general-pde} with
\begin{equation} \label{eq: burgerPDE}
    g \left(x, \frac{\partial x}{\partial \eta}, \frac{\partial^2 x}{\partial\eta^2}; \mu \right)  
    = 
    \mu \frac{\partial^2}{\partial \eta^2} x(\eta, t) 
    - x(\eta, t) \frac{\partial}{\partial\eta}x(\eta,t)  
\end{equation}
where $x(\eta, t)$ is the velocity of the fluid, $t$ is the time, $\eta$ is the spatial coordinate, and $\mu$ is the kinematic viscosity coefficient. Equation \eqref{eq: burgerPDE} can be derived from the Navier-Stokes equations by assuming one-dimensional flow and neglecting the effects of pressure.

\subsubsection{Setup}
We define the domain and parameters of our Burgers equation test problem as follows. Let $\Omega = [0, L] \subset \real$ where $L = 1$ be the spatial domain and $t \geq 0$ the time domain. Let parameter $\mu = 0.01$. Discretization of the spatial domain $\Omega$ of \eqref{eq: burgerPDE} with $N=128$ uniformly-spaced points using the finite difference method yields \eqref{eq: full_lq_model}. We consider the initial condition described in \eqref{eq:kl-expansion}, and let the Gaussian random field parameters be $\tau = 1$ and $s = 3$. We test reduced dimensions $r = 1, \dots, 12$ in our experiments which corresponds to retaining about $90$\% of the energy according to \Cref{fig:energy_loss}. To collect training data, we draw coefficients $\{\xi_i\}_{i = 1}^{N}$ \gls{iid} from $\mathcal{N}(0, 1)$ for each initial condition. 

High-fidelity time derivative data $\{\xdot_k\}$ arises from simulating \eqref{eq: full_lq_model} for the initial condition. The low-fidelity model arises from discretizing the spatial domain $\Omega$ of \eqref{eq: burgerPDE} into $N_\ell = 16$ uniformly-spaced points using the finite difference method, yielding \eqref{eq: low_lq_model}.

\begin{figure}[b!]
    \centering
    \includegraphics[width=\linewidth]{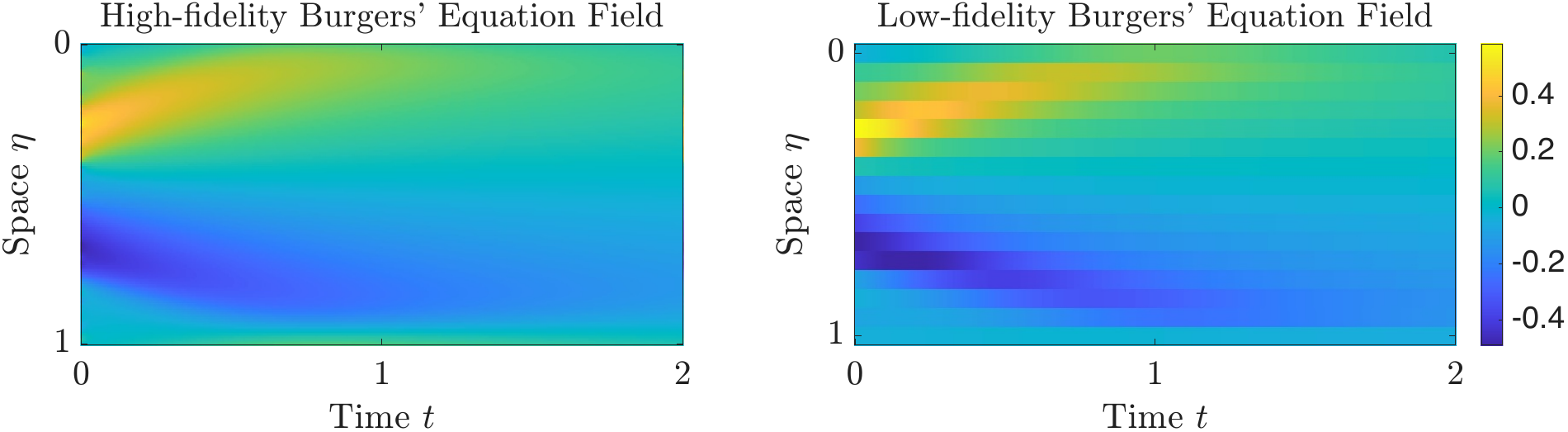}
    \caption{Visualization of high versus low-fidelity discretization of the 1D Burgers' Equation finite difference model, integrated to final time $t = 2$ with $\Delta t = 10^{-3}$. A random initial condition is drawn from training data set as described in the experimental setup.}
    \label{fig: flow_fields}
\end{figure}

The low-fidelity data $\{\dot{\tilde{\xvec}}_k\}$ is collected by simulating \eqref{eq: low_lq_model} for the initial condition. A visualization of the high and low-fidelity discretization of a solution field integrated using a semi-implicit Euler scheme to a final time of $t = 2$ with $\Delta t = 10^{-3}$ is provided in \Cref{fig: flow_fields}.

The POD basis $\mathbf{\Phi}_r$ is obtained analytically from the \gls{grf} covariance, and we reduce high and low-fidelity data to obtain $\{\yvec^{(1)}_k\}$ and $\{\yvec^{(2)}_k\}$ respectively as demonstrated in \eqref{eq: hf_model} and \eqref{eq: lf_model}. The paired feature transformations $\{\dvec_k\}$ are formed using \eqref{eq: opinf_feature}. Model statistics for hyperparameter selection as discussed in \cref{subsection: hyperparameter} are calculated assuming perfect statistics are available. Statistics are obtained by generating a large dataset of $3 \times 10^6$ snapshots.

The standard \gls{opinf} \eqref{eq: hfopinf_decoupled} and \gls{mfopinf} \eqref{eq: mfopinf_decoupled} estimators for $\ohat_i^*$ are computed by drawing \gls{iid} snapshots from the \gls{grf} within budget constraint $p$. We normalize the model costs so that the cost to evaluate one high-fidelity snapshot has cost $w_1 = 1.00$ and the cost to evaluate one low-fidelity snapshot has cost $w_2 = 1.56 \times 10^{-2}$, where costs are calculated from \gls{flops} needed to evaluate a snapshot.

We train $M_\text{rep} = 100$ replicates of \gls{opinf} operators and \gls{mfopinf} operators, where replicates are trained with different realizations of random training data. To determine the sample size $K$ for training single, high-fidelity \gls{opinf} operators, we divided the budget by the high-fidelity cost $K = p/w_1$. Sample allocation for \gls{mfopinf} $K_{1}, K_{2}$ is determined using \eqref{eq: sampleallocation}, and the scaling coefficient $\{\alpha_i\}_{i = 1}^r$ is determined using \eqref{eq: optimalalpha}. 

We generate testing data with $M_\text{test} = 20$ \gls{iid} drawn initial conditions from the training region as described in \eqref{eq:kl-expansion}. The system of \gls{ode}s \eqref{eq: full_lq_model} is marched forward in time for each new initial condition with a semi-implicit Euler scheme at timestep size $\Delta t = 10^{-3}$ with final time $t = 2$ to generate a test dataset of $M_\text{test}$ flow fields $\{\Xmat_i\}_{i=1}^{M_\text{test}}$. We use the learned models to reconstruct the additional data by simulating them from the same new initial conditions $\{\mathbf{\Phi}_r \Xhat_{j, i}\}_{i=1}^{M_\text{reps}}$, where $\Xhat_{j, i}$ is the predicted reduced-dimensional flow field from the $j^\text{th}$ operator replicate and $i^\text{th}$ test initial condition. We report the averaged state error of the reduced model reconstruction defined as using \eqref{eq: state_error}. The averaged state error is calculated for $\Xhat_{j, i}$ obtained from both single, high-fidelity trained \gls{opinf} models and \gls{mfopinf} trained models, for reduced dimension sizes $r = 1, \dots, 12$, and for budget sizes $p =  1000$, $10000$, and $100000$.

\subsubsection{Results}

\begin{figure}[t]
    \centering
    \includegraphics[width=\linewidth]{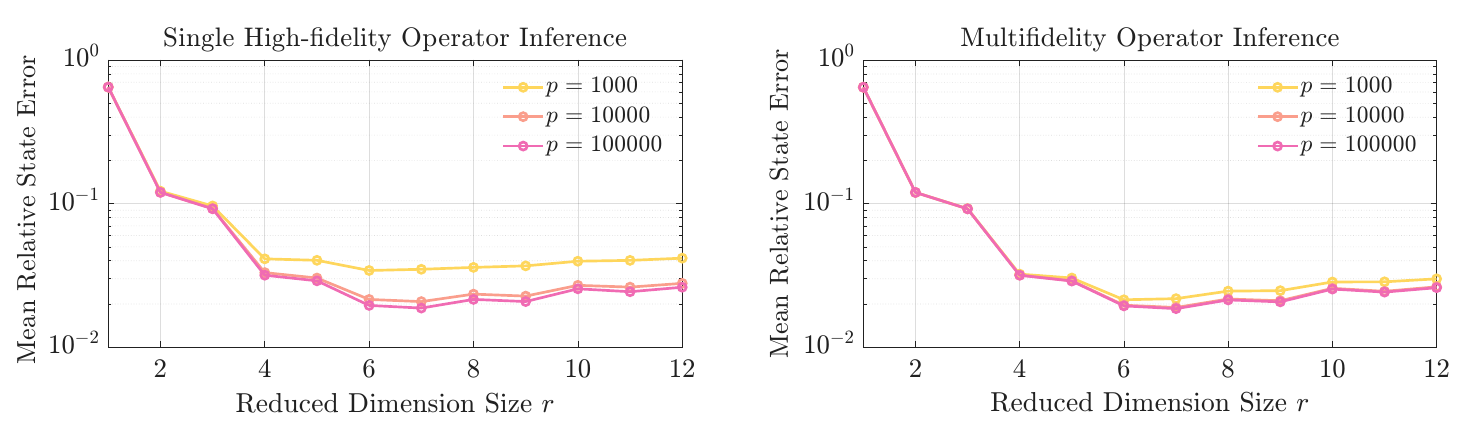}
    \caption{Comparison of mean relative state error between the standard single, high-fidelity \gls{opinf} (left) and the new \gls{mfopinf} field predictions for the 1D Burgers' equation example}
    \label{fig: burgstateerror}
\end{figure}

\Cref{fig: burgstateerror} shows the average relative state error over trained model replicates and over different test fields for different training budgets and reduced dimension size for single, high-fidelity \gls{opinf} (left) and \gls{mfopinf} (right). Across all reduced dimensions $r$, we observe a lower mean relative state error for the \gls{mfopinf} results compared to the single, high-fidelity \gls{opinf}. The difference in performance is most prominent in the lower budget sizes where high-fidelity data is scarce, which demonstrates how low-fidelity data can compensate for the lack of high-fidelity data.

\begin{figure}[t]
    \centering
    \includegraphics[width=0.7\linewidth]{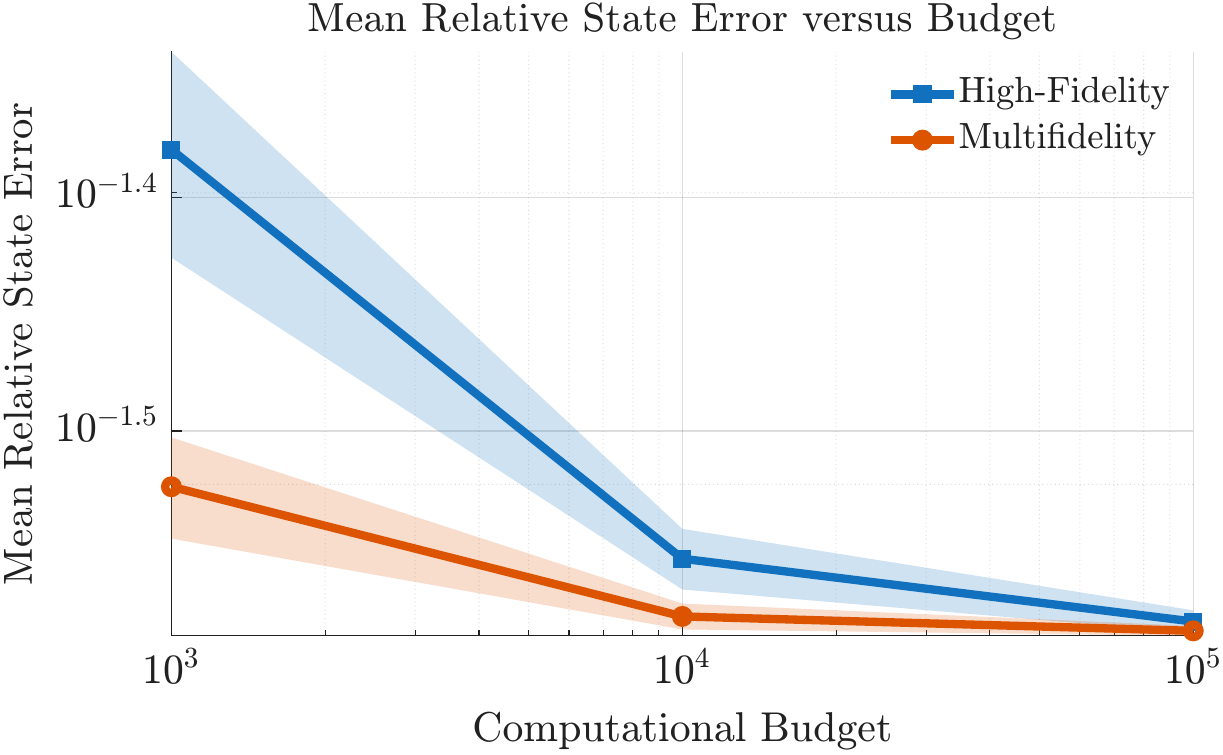}
    \caption{Comparison of mean relative state error $\pm$ one standard deviation between the standard single, high-fidelity \gls{opinf} and \gls{mfopinf} learned with reduced dimension size $r = 12$ for the 1D Burgers' Equation example.}
    \label{fig: burgstateerror_12}
\end{figure}

In \Cref{fig: burgstateerror_12}, we plot the mean relative state error for just reduced dimension size $r = 12$ with respect to budget sizes $p = 1000, 10000, 100000$. For low budgets, we observe the greatest benefit in implementing \gls{mfopinf} over the standard \gls{opinf}, again demonstrating the efficacy of utilizing a multifidelity approach when high-fidelity data is scarce.

Similar to the heat equation results, we note that in both \Cref{fig: burgstateerror} and \Cref{fig: burgstateerror_12} the mean relative state errors are higher orders of magnitude than standard model reduction/\gls{opinf} literature which may be attributed to extrapolation past the training regime formed by initial conditions.

\subsection{Kuramoto-Sivashinsky Equation} \label{subsection: KSE}
In this numerical example, we consider the Kuramoto-Sivashinsky equation (KSE), a \gls{pde} that was originally discovered to describe the dynamics of flame propagation and has since been extended to fluid dynamics, pattern formation, and combustion theory \cite{kuramoto, sivashinsky}. The \gls{pde} is
\begin{equation} \label{eq: KSPDE}
    \frac{\partial }{\partial t}x(\eta, t) = - \frac{\partial^2 }{\partial \eta^2}x(\eta, t) - \mu\frac{\partial^4 }{\partial \eta^4}x(\eta, t) - x(\eta, t)\frac{\partial }{\partial \eta} x(\eta, t)
\end{equation}
where $x(\eta, t)$ is the velocity of the fluid, $t$ is the time, $\eta$ is the spatial coordinate, and $\mu$ is the kinematic viscosity coefficient. 

The chaotic nature of the \gls{kse} means that small perturbations in the current state can lead to large deviations in later states. In particular, the \gls{kse} exhibits a phenomenon called \textit{correlation splitting}. Correlation splitting is a key indicator of chaos and indicates that after a certain time $t > t^*$, the system state is statistically independent of its past and future states \cite{dynamical_chaos_correlation, KS_complexity}.

\subsubsection{Setup}
We define the domain and parameters of the \gls{kse} test problem following~\cite{koike2024energy,koike2026streaming}. Let $\Omega = [0, L] \subset \real$ where $L = 22$ be the spatial domain and $t \geq 0$ the time domain. Let parameter $\mu = 1$. Discretization of the spatial domain $\Omega$ of \eqref{eq:  KSPDE} with $N=512$ uniformly-spaced points using the finite difference method yields \eqref{eq: full_lq_model}. We consider initial conditions of the form
\begin{equation} \label{eq: KS_IC}
    x_0(\eta) = a_0\cos \left(\frac{2\pi \eta}{L} \right) + b_0\sin\left(\frac{4\pi \eta}{L}\right)
\end{equation}
where $a_0$ and $b_0$ are parameters for the initial condition. 

Unlike the numerical experiments in \cref{subsection: heateq} and \cref{subsection: burgeq}, we do not need to generate a dataset from initial conditions described in \cref{subsection: ICs} because of the correlation splitting phenomenon exhibited by the \gls{kse}. The statistical independence of states at $t>100$ may be interpreted as random samples of snapshots from a distribution $\pi_0$. To collect training data, we consider parameters $a_0 \in \{ 0.8, 1.0, 1.2, 1.4 \}$ and $b_0 \in \{ 0.2, 0.4, 0.6\}$ for twelve initial conditions and trajectories. Then, we collect every $100^\text{th}$ snapshot from the high and low-fidelity model at $100 < t < 300$. We repeat this process to generate a dataset of $2.4 \times 10^4$ random state snapshots.

High-fidelity time derivative data $\{\xdot_k\}$ arises from simulating \eqref{eq: full_lq_model} using an implicit Crank-Nicolson and Adams-Bashforth scheme for one timestep at size $\Delta t = 10^{-3}$. The low-fidelity model arises from discretizing the spatial domain $\Omega$ of \eqref{eq: KSPDE} into $N_\ell = 64$ uniformly-spaced points using the finite difference method, yielding \eqref{eq: low_lq_model}.

\begin{figure}[H]
    \centering
    \includegraphics[width=\linewidth]{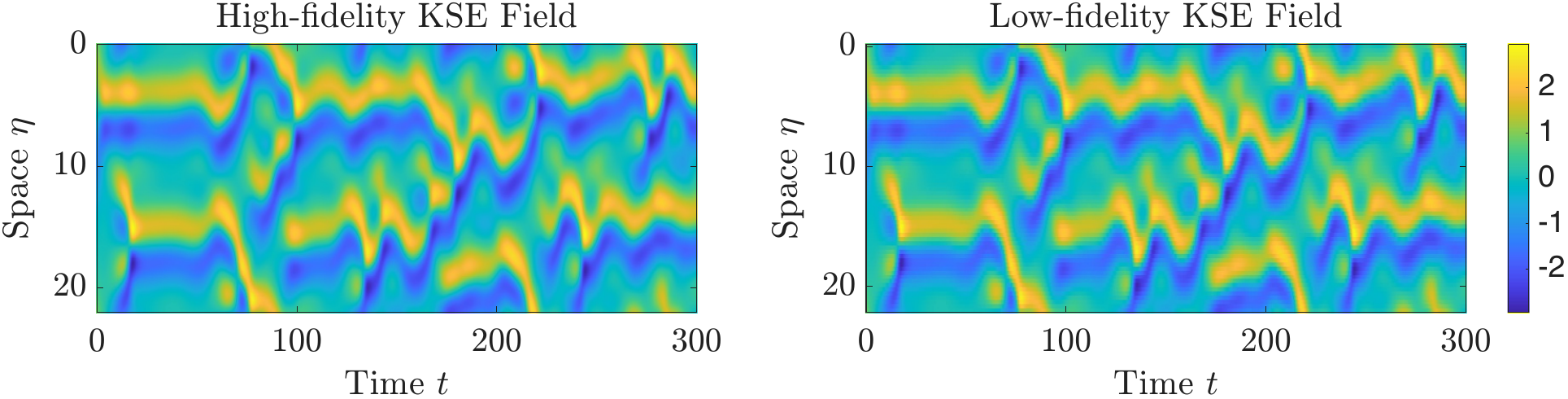}
    \caption{Visualization of the high versus low-fidelity discretization of the 1D Kuramoto-Sivashinsky model, using initial condition $x_0 (\eta)= 0.9\cos(2\pi \eta/L) + 0.3 \sin (4\pi \eta/L)$.}
    \label{fig:flow_field_ks}
\end{figure}

The low-fidelity data $\{\dot{\tilde{\xvec}}_k\}$ is collected by simulating \eqref{eq: low_lq_model} using an implicit Crank-Nicolson and Adams-Bashforth scheme for one timestep at size $\Delta t_\ell = 10^{-2}$. A visualization of the high and low-fidelity discretization of a solution field integrated to a final time of $t = 300$ is provided in \Cref{fig:flow_field_ks}.

\begin{figure}[H]
    \centering
    \includegraphics[width=0.6\linewidth]{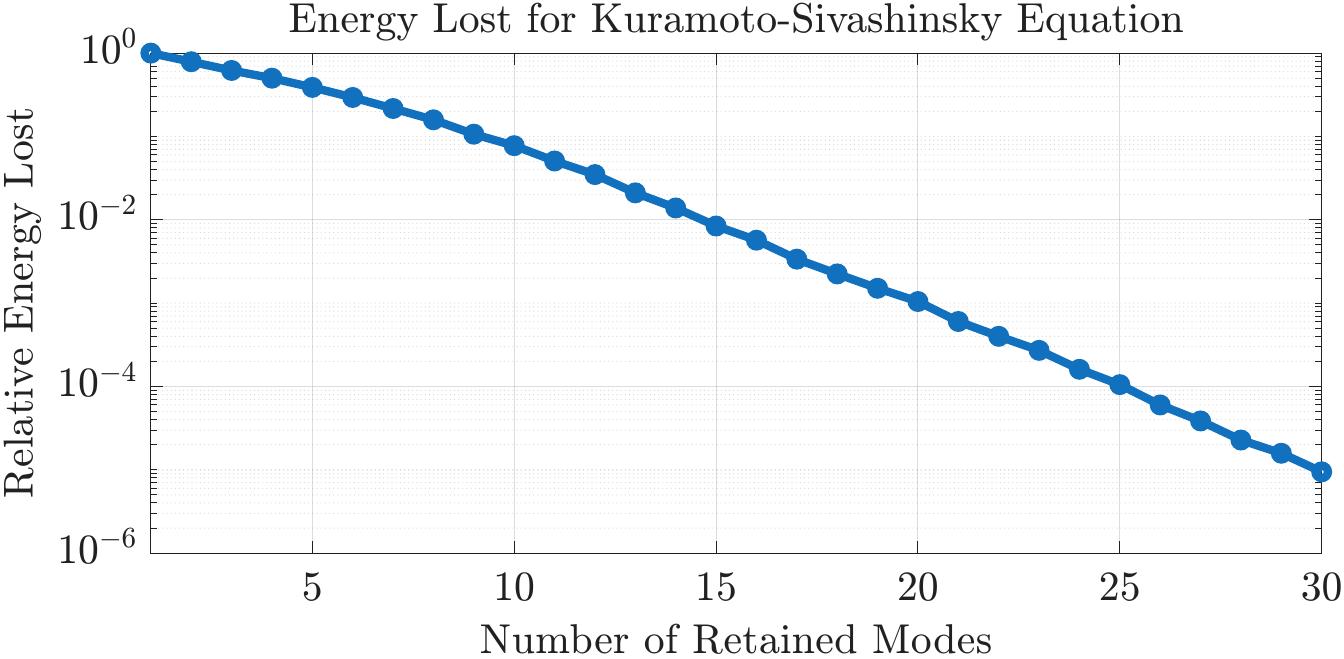}
    \caption{Relative energy loss from truncating to $r$ modes for the Kuramoto-Sivashinsky Equation Example}
    \label{fig: KS_energy_loss}
\end{figure}

The \gls{pod} basis $\mathbf{\Phi}_r$ is obtained using all $2.4 \times 10^4$ state snapshots samples from the high-fidelity dataset. To compute the relative energy lost as a function of the reduced dimension or the number of retained modes $r$, we use
\begin{equation}
    \text{Relative Energy Lost:} \quad \frac{\sum_{i = r + 1}^N\sigma^2_i}{\sum_{i =1}^N\sigma^2_i}
\end{equation}
where $\sigma_i$ comes from the singular values of the state snapshots $\Sigmamat = \text{diag}(\sigma_1, \dots, \sigma_N)$. The computed singular values from all $2.4 \times 10^4$ snapshots are shown in \Cref{fig: KS_energy_loss}.

We note that when we retain $r_\text{max} = 12$ modes, we retain about $99$\% of the energy in the training data. Thus, we test reduced dimension sizes $r = 1,\dots, 12$ in our experiments. Using the \gls{pod} basis $\mathbf{\Phi}_r$, we reduce high and low-fidelity data to obtain $\{\yvec^{(1)}_k\}$ and $\{\yvec^{(2)}_k\}$ respectively as demonstrated in \eqref{eq: hf_model} and \eqref{eq: lf_model}. The paired feature transformations $\{\dvec_k\}$ are formed using \eqref{eq: opinf_feature}. Model statistics for hyperparameter selection as discussed in \cref{subsection: hyperparameter} are calculated using all $2.4 \times 10^4$ available samples.

The standard \gls{opinf} \eqref{eq: hfopinf_decoupled} and \gls{mfopinf} \eqref{eq: mfopinf_decoupled} estimators for $\ohat_i^*$ are computed by bootstrapping samples from this dataset within budget constraint $p$. We assume that the cost to evaluate one high-fidelity snapshot has cost $w_1 = 1.00$ and the cost to evaluate one low-fidelity snapshot has cost $w_2 = 1.56 \times 10^{-2}$, where costs are calculated from \gls{flops} needed to evaluate a snapshot.

We train $M_\text{rep} = 100$ replicates of \gls{opinf} operators and \gls{mfopinf} operators, where replicates are trained with different realizations of random training data from bootstrapping from the dataset. To determine sample size $K$ for training single, high-fidelity \gls{opinf} operators, we divided the budget by the high-fidelity cost $K = p/w_1$. Sample allocation for \gls{mfopinf} $K_1, K_2$ is determined using \eqref{eq: sampleallocation}, and the scaling coefficients $\{\alpha_i\}_{i = 1}^r$ are determined using \eqref{eq: optimalalpha}. 


The chaotic nature of the \gls{kse} demonstrates that small perturbations in the state, e.g., in the initial condition, can lead to large differences in state at a later time. Thus, the standard state error metric used in the previous two examples are not suitable for the \gls{kse} example. Rather than comparing models with their point-wise predictions, we report the accuracy of the reduced operator as follows:
\begin{equation}\label{eq: operator_error}
    \text{Mean Relative Operator Error: } \frac{1}{M_\text{reps}} \sum_{j = 1}^{M_{\text{reps}}}\frac{\norm{\Omat^*-\Ohat_j}^2_\text{F}}{\norm{\Omat^*}^2_\text{F}}
\end{equation}
where $\Omat^*$ is taken to be the true reduced operator, calculated with all $2.4 \times 10^4$ snapshots from the dataset, and $\Ohat_j$ is the reduced operator obtained from both single, high-fidelity trained \gls{opinf} models and \gls{mfopinf} trained models, for reduced dimension sizes $r = 1, \dots, 12$, and for budget sizes $p = 1000, 2000, 5000, 10000$.

\subsubsection{Results}
\begin{figure}[H]
    \centering
    \includegraphics[width=1\linewidth]{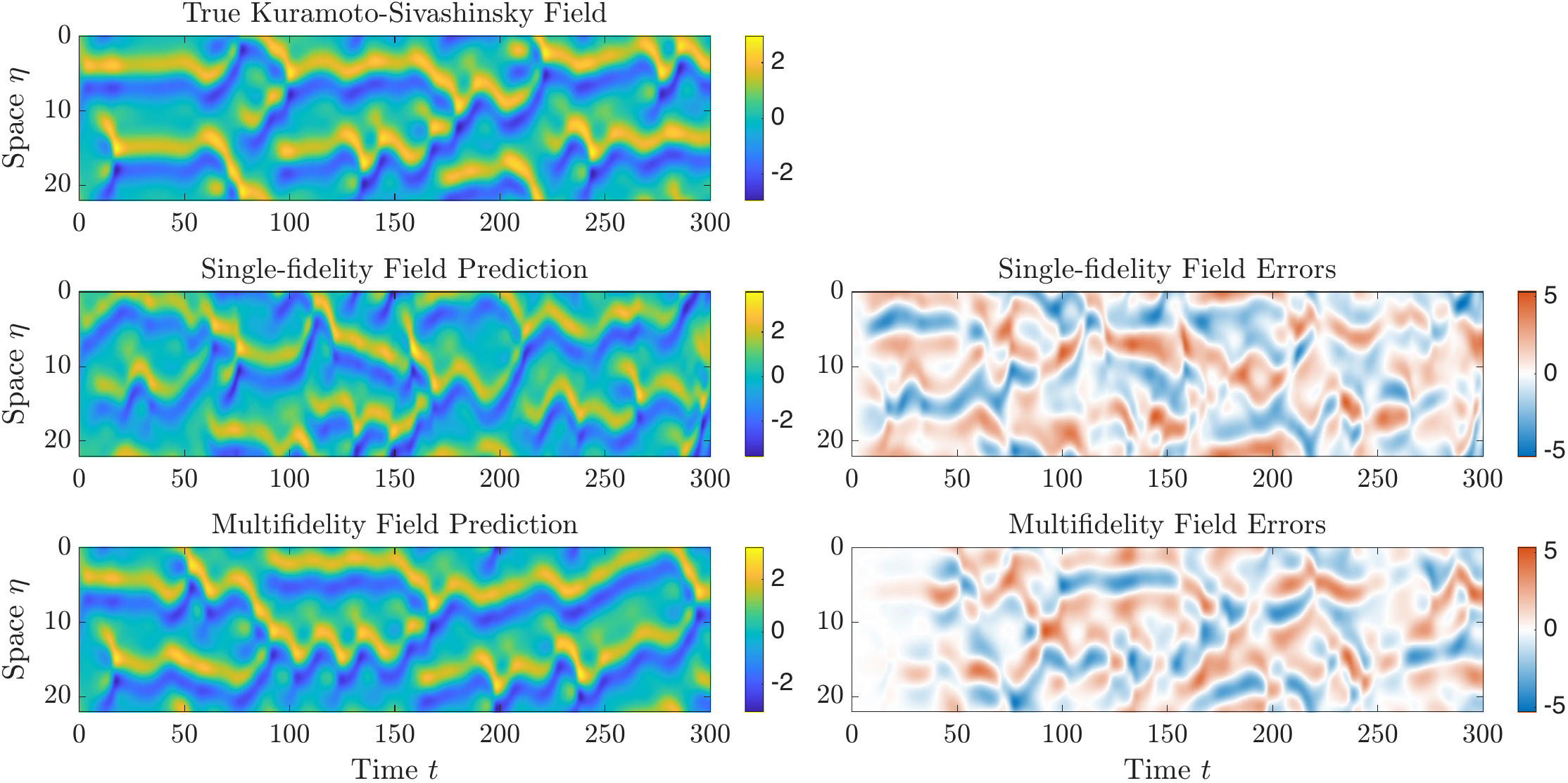}
    \caption{Point-wise errors between one replicate of the standard single, high-fidelity \gls{opinf} and \gls{mfopinf} for the 1D Kuramoto-Sivashinsky example with $r= 12$.}
    \label{fig: field_error}
\end{figure}

\Cref{fig: field_error} shows the point-wise errors between a single, high-fidelity trained \gls{opinf} and \gls{mfopinf} operator using one budget size of $15000$. This plot uses a test initial condition with parameters $a_0 = 0.9$ and $b_0 = 0.3$ and compares the reduced operators to the full order model. All trajectories are integrated to a final time of $T = 300$ with $\Delta t = 10^{-3}$ on a domain $\eta \in \Omega = [0, 22]$ with $512$ uniformly-spaced points. While we observe lower magnitude errors for the multifidelity field prediction when $t < 50$, we notice high magnitudes of error due to the chaotic nature of the \gls{kse}. Thus, operator error will be the main metric for comparison.

\begin{figure}[H]
    \centering
    \includegraphics[width=1\linewidth]{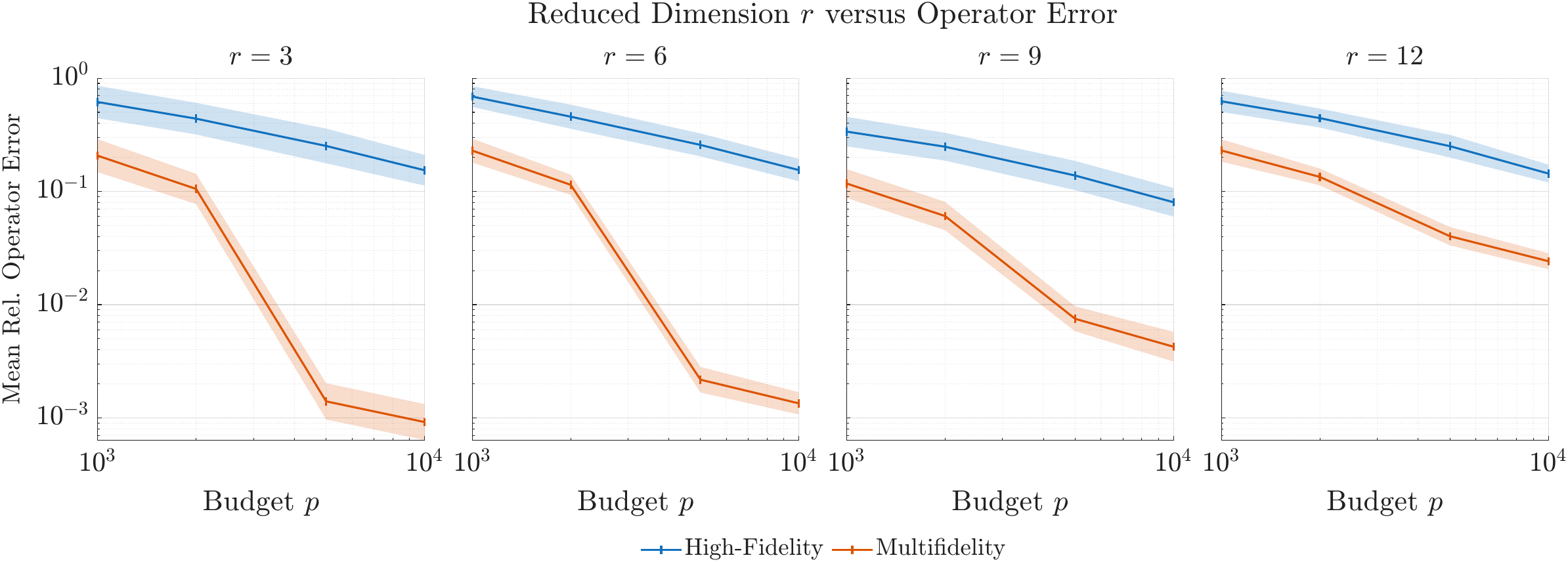}
    \caption{Mean relative operator error for reduced dimension sizes $r = 3, 6, 9, 12$ between the standard single, high-fidelity \gls{opinf} and \gls{mfopinf} for the 1D Kuramoto-Sivashinsky example. Bounds are $\pm$ one standard deviation of $100$ operator replicates.}
    \label{fig: KSE_op_err}
\end{figure}

\Cref{fig: KSE_op_err} shows the mean relative operator error over trained model replicates for reduced dimension sizes $r = 3, 6, 9, 12$ versus the budget. Across all reduced dimensions, we observe a lower mean relative operator error for the \gls{mfopinf} results compared to the single, high-fidelity \gls{opinf}. The largest increase in error reduction occurs at smaller reduced dimension sizes, and the benefit of \gls{mfopinf} appears to decrease as the reduced dimension size increases. This phenomenon may be attributed to a loss of correlation between high and low-fidelity data when more modes are retained. 

\begin{figure}[H]
    \centering
    \includegraphics[width=1\linewidth]{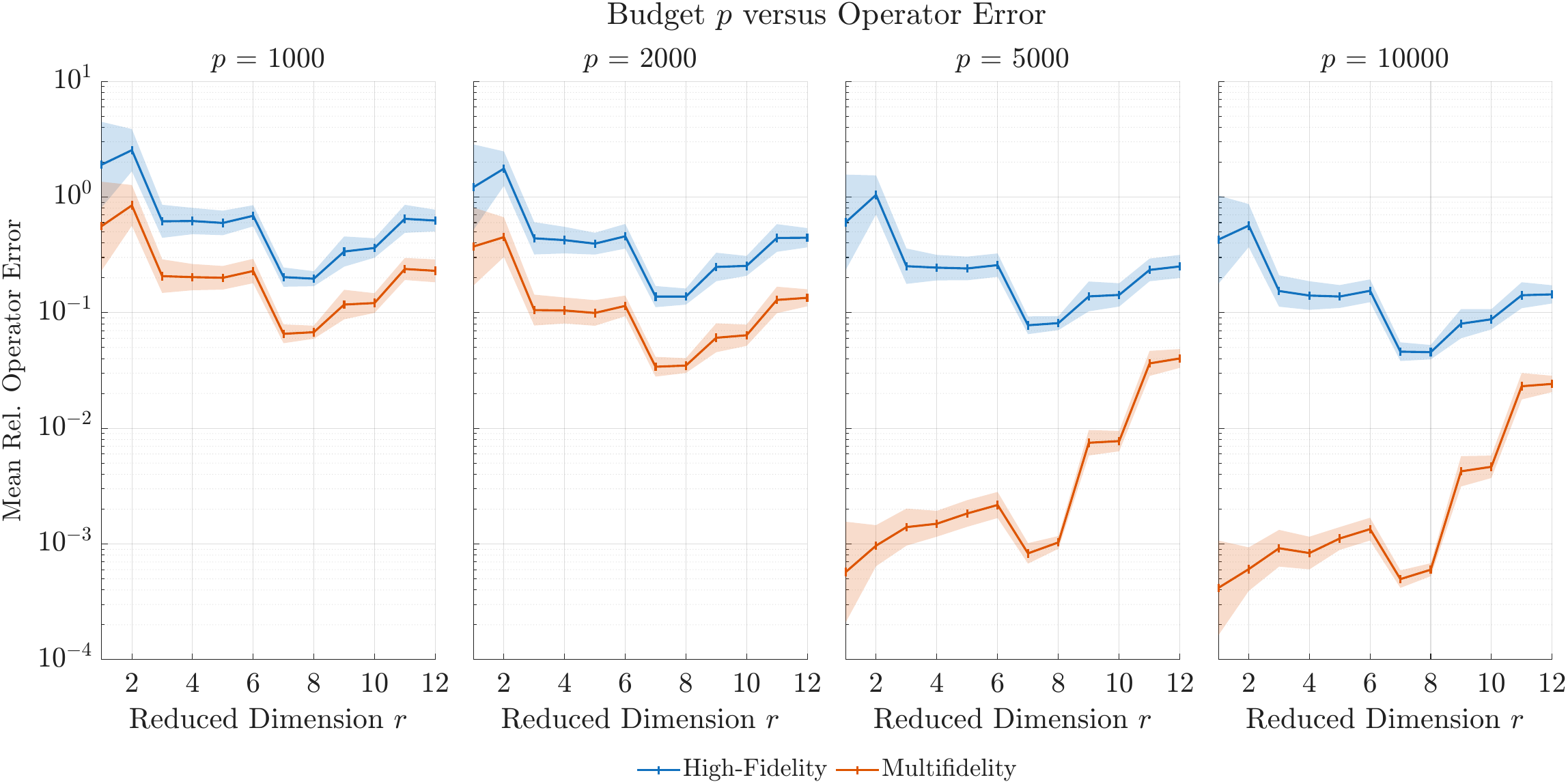}
    \caption{Comparison of the mean relative operator error between the standard single, high-fidelity \gls{opinf} and \gls{mfopinf} for budgets $p = 1000, 2000, 5000, 10000$ and $r = 1, \dots, 12$.}
    \label{fig: budget_kse}
\end{figure}

In \Cref{fig: budget_kse}, we plot the mean relative operator error across reduced dimension sizes $r = 1, \dots, 12$ with respect to budget sizes $p = 1000, 2000, 5000, 10000$. Across subplots, we observe the same trend as \Cref{fig: KSE_op_err} where for a fixed reduced dimension $r$, larger budgets exhibit smaller operator errors. There is a significant decrease in error for the multifidelity trained operators compared to the single high-fidelity trained operators for the larger budget sizes tested.

Due to plotting the operator errors across dimensions on the same axis, we may attribute increasing error trends at $r = 7$ to an increase in dimensionality. However, we still observe lower operator errors for \gls{mfopinf} compared to single, high-fidelity \gls{opinf} across all budgets and reduced dimension sizes. Again, we observe the largest increase in error reduction for smaller reduced dimension sizes for $p = 5000$ and $10000$ which may be attributed to less correlation between high and low-fidelity data at larger reduced-dimension sizes.

\section{Conclusion} \label{section: conclude}
We present a new multifidelity approach to \gls{opinf} for learning high-dimensional dynamical systems. Often when learning models from scientific or engineering data, high-fidelity data is limited, leading to models with high variance or error. Our approach leverages low-fidelity data to improve the robustness of operators trained on scarce high-fidelity data. Utilizing a control variate framework for linear regression parameter estimation, we are able to construct a column-wise, unbiased estimator for the reduced operators in the \gls{opinf} model that utilizes both high and low-fidelity data. 

We provide results from three numerical experiments: the 1D heat equation, 1D viscous Burgers' equation, and the Kuramoto-Sivashinsky equation (KSE). All numerical examples demonstrate that the new \gls{mfopinf} approach reduces the variance, and therefore, the generalization error of the learned model relative to single-fidelity training on high-fidelity data only. This effect is most pronounced at low training budgets, i.e., scarce high-fidelity data for the heat and Burgers' equation examples. These results support the multifidelity approach, even when extrapolating past the initial timestep from training data. For the \gls{kse} example, we observe this effect for the higher budgets tested. For further experimentation, we will expand the budgets tested for the \gls{kse} example.

We impose optimal scalar control variate coefficient and sample allocation for column-by-column estimation of the reduced operator. Some directions for further exploration include matrix coefficients for a unified estimation of the reduced operator and budget allocation for snapshot data collection. Additionally, the choice of the \gls{pod} basis heavily impacts the performance of \gls{opinf} when predicting full-dimensional fields. For future numerical experiments, we hope to explore other choices for initial condition distributions, e.g., with different eigenvalue decay or different eigenvalue hyperparameter selection. We may also consider different methods of obtaining the \gls{pod} basis. Currently, the \gls{pod} basis is obtained analytically from a known distribution, or from a sample of high-fidelity data. Future work may explore the data-driven method of obtaining the \gls{pod} basis from multifidelity data. 

Directions for future work include extensions for learning to more complex, turbulent models with scarce high-fidelity data, learning from different levels of low-fidelity data, or alternate control variate frameworks for multifidelity estimation of operators. This includes studying different control variate frameworks (MLMC, MLBLUE, etc.) or direct estimation of the reduced operator or model rather than column-by-column. 

\section*{Acknowledgments}

VZ, TK, AR, and EQ were supported in part by the National Science Foundation under CMMI-2442140. Additionally, VZ, TK, and EQ were supported by the US Department of Energy Office of Science Energy Earthshot Initiative as part of the ‘Learning reduced models under extreme data conditions for design and rapid decision-making in complex systems’ project under award number DE-SC0024721. Any opinions, findings, and conclusions or recommendations expressed in this material are those of the authors and do not necessarily reflect the views of the National Science Foundation or the US Department of Energy.

\bibliography{ref}

\end{document}